\documentclass[twocolumn,10pt]{asme2ej}

\usepackage{graphicx,enumerate}
\usepackage{subcaption}
\usepackage{epsfig}  
\usepackage{amssymb,amsmath} 
\usepackage[all]{xy}
\usepackage{multicol}
\usepackage{xcolor}
\definecolor{aliceblue}{rgb}{0.94, 0.97, 1.0}
\definecolor{green}{rgb}{0.0, 0.5, 0.0} 
\usepackage{booktabs,chemformula}

\usepackage{float}
\usepackage{afterpage}
  
\usepackage{mathptmx}       
\usepackage{helvet,overpic}         
\usepackage{courier}        
\usepackage[bottom]{footmisc} 
\usepackage{amsmath,amssymb,epsfig} 
      
\newtheorem{rmk}{Remark}

\usepackage{hyperref}

\title{Architecture Singularity Distance Computations \\
for Linear Pentapods}

\author{Aditya Kapilavai and Georg Nawratil
    \affiliation{
	Institute of Discrete Mathematics and Geometry\\
	Vienna University of Technology\\
	Wiedner Hauptstrasse 8-10/104, Vienna 1040, Austria\\
  Email:{\{akapilavai,nawratil\}@geometrie.tuwien.ac.at}
    }	  
}

\begin{document}

\maketitle    

\begin{abstract}
{\it
The kinematic/robotic community is not only interested in measuring the closeness of a given robot configuration to its next singular one 
but also in a geometric meaningful index evaluating how far the robot design is away from being architecturally singular. Such an architecture singularity distance, which can be used by 
engineers as a criterion within the design process, is presented for 
a certain class of parallel manipulators of Stewart-Gough type; namely so-called linear pentapods. 
Geometrically the architecture singular designs are well-understood and can be subclassified into several cases, which allows to solve the optimization problem of 
computing the closest architecture singular design to a given linear pentapod with algorithms from numerical algebraic geometry.}
\end{abstract}

\section{Introduction}\label{intro}

A linear pentapod is a five-degree-of-freedom parallel manipulator of the Stewart-Gough type, where the linear motion platform is linked via five SPS legs to the base.  
Note that only the prismatic (P) joints are actuated and all spherical (S) joints are passive. 
The geometry of a linear pentapod is given by the five base anchor points with coordinates $\mathbf{M}_{i}:=(x_i,y_i,z_i)_{\mathfrak{F}_{0}}$ with respect to the 
fixed frame $\mathfrak{F}_{0}$ and by the five collinear platform anchor points $\mathbf{m}_{i}:=(r_i,0,0)_{\mathfrak{F}}$ with respect to the moving frame  $\mathfrak{F}$ (for $i=1,\ldots,5$). 
Industrial applications of linear pentapod are e.g.\ discussed in  \cite{b0} and \cite{b1}.

\begin{figure}[ht]
\begin{center}
\begin{overpic}[width=6cm]{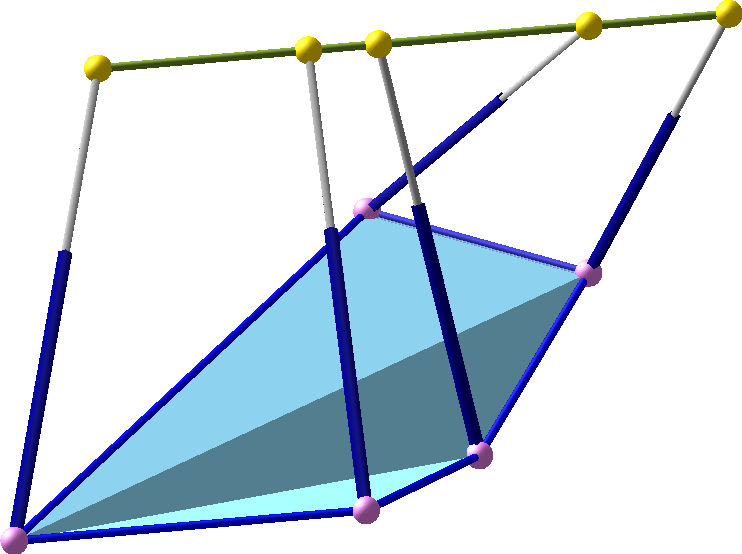}
\begin{small}
\put(0,-6){$\mathbf{M}_{1}$}
\put(-3,2.9){S}
\put(30,10){{Base}}
 \put(34,18){$\mathfrak{F}_0$}
\put(-1,20){\textcolor{blue}{P}}
\put(7,60){S}
\put(15,60){{Platform}}
\put(45,-2){$\mathbf{M}_{2}$}
\put(60,5){$\mathbf{M}_{3}$}
\put(80,31){$\mathbf{M}_{4}$}
\put(44,51){$\mathbf{M}_{5}$}
\put(10,68){$\mathbf{m}_{1}$}
\put(35.5,72){$\mathbf{m}_{2}$}
\put(48,72){$\mathbf{m}_{3}$}
\put(72,76){$\mathbf{m}_{5}$}
\put(92,77){$\mathbf{m}_{4}$}
 \put(27,70){$\mathfrak{F}$}
\end{small}
\end{overpic}
\end{center}
\caption{Schematic representation of a linear pentapod.}
\label{5SPS}
\end{figure}
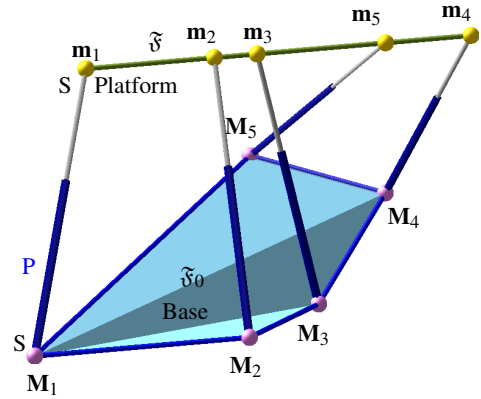

From the line-geometric point of view, a linear pentapod is in a singular configuration if and only if the Pl\"ucker coordinates of the carrier lines of the five legs 
are linearly dependent.
Linear pentapods, which are singular in every possible configuration, are called architecture singular \cite{b2}.
It is well-known \cite{b4} that these manipulators are redundant and therefore 
cannot be controlled. 
As every anchor point can be associated with a space of uncertainties resulting from tolerances in the S-joints and deviations 
of the platform and the base from the geometric model as well as their deformations under operation, not only architecture singularities 
should be avoided but also their vicinity.
Therefore, there is a need for a metric allowing the evaluation of the distance of a given design to the closest architecture singularity.

\subsection{Review}\label{sec:review}

Borras et al presented in \cite{b3} an index evaluating how far a given linear pentapod with a planar base, where no platform or base points coincide, is away from being architecturally singular. This index is based on the geometric property that there exists a bijection between the points of the linear platform and the lines of a pencil located in the base plane. In the assumed generic case the pencil vertex $\mathbf{B}=(b_{x}, b_{y},0)_{\mathfrak{F}_{0}}$ differs from the five base points. For this case, the architectural singularity can be characterized by the geometric condition that the six-points $\mathbf{M}_1,\ldots,\mathbf{M}_5,\mathbf{B}$ are located on a conic section, which is characterized by the vanishing of the following determinant:
\begin{equation}
C=
   \begin{vmatrix}
     b_{x}^{2} & b_{x}b_{y} &  b_{y}^{2} & b_{x} & b_{y} & 1\\ 
     x_{1}^2 & x_{1}y_{1} & y_{1}^2  & x_{1} & y_{1} & 1 \\
     \vdots & \vdots  & \vdots & \vdots & \vdots & \vdots \\
     x_{5}^2 & x_{5}y_{5} & y_{5}^2  & x_{5} & y_{5} & 1
\end{vmatrix}
\label{borrasindex}
\end{equation}

If this is not the case the obtained value for $C$ {\it provides a useful index to be optimized in the design process, to obtain manipulators as far as possible from architectural singularities} according to \cite{b3}.

The limitations of this approach are as follows: 
Firstly, it can only be applied to generic linear pentapods with a planar base\footnote{If $\mathbf{B}$ coincides with one of the points $\mathbf{M}_1,\ldots,\mathbf{M}_5$ then also $C=0$ holds but the manipulator has not to be architecturally singular.}. 
Secondly, the value computed in  Eq.~(\ref{borrasindex}) is not a distance but only an index.   

\subsection{Outline}
In Section \ref{sec:dist}, we present such a distance function, and in Section \ref{distances}, it is discussed how the corresponding optimization task can be managed by breaking it up into several minimization problems, which are related to the different classes of architectural singularity known in the literature \cite{b4}. This fragmentation of the computation reduces the maximal number of needed unknowns, allowing a solution with tools of numerical algebraic geometry using the software \texttt{Bertini} \cite{b6} and \texttt{HC.jl} \cite{b11}, respectively. 
The developed computational pipeline for computing the architectural singularity distance is discussed in Section~\ref{sec:procedure1}.  In Section~\ref{results}, we present two numerical examples, where the first design has a planar base and the second manipulator has a non-planar one. 
The first example is also compared with the approach pointed out in Section~\ref{sec:review}. We conclude the paper in Section \ref{future}.

\section{Architecture singularity distance function}\label{sec:dist}

In \cite{b5} a metric was presented for the evaluation of the extrinsic distance between two configurations of linear pentapods. 
One can use a similar idea to \cite{b5} for the definition of an architectural singularity distance $D$ with
\begin{equation}
D^{2} := {\frac{1}{10}\sum_{i=1}^{5}\left[||{\mathbf{M}^{'}_{i}}-\mathbf{M}_{i}||^{2}+||{\mathbf{m}^{'}_{i}}-\mathbf{m}_{i}||^{2}\right]},
\label{eq:distance0}
\end{equation} 
where $\mathbf {M}^{'}_{i}:=(x^{'}_{i},y^{'}_{i},z^{'}_{i})_{\mathfrak{F}_0}$  and $\mathbf{m}^{'}_{i}:=(r^{'}_{i},0,0)_{\mathfrak{F}}$ denote the base and platform anchor points of the closest 
architecture singular design. Note that Eq.\  (\ref{eq:distance0}) only depends on the geometry of the linear pentapod and not on the relative pose of the 
platform to the base, which is the case for the metric of \cite{b5}. 

In analogy to \cite[Thm.\ 1]{b5}, we can also give a physical interpretation of the architecture singular distance $D$ of  Eq.\  (\ref{eq:distance0}), which reads as follows: 
Let $\Omega_i$ (resp.\ $\omega_i$) be the radius of the smallest $2$-sphere (resp.\ $0$-sphere)  centered in the base (resp.\ platform) anchor point of the geometric model, 
which contains the associated space of uncertainties. 
If $max(\Omega_1,\ldots,\Omega_5,\omega_1,\ldots,\omega_5)<D$ holds then the mechanical design of the linear pentapod is guaranteed to be not architecturally singular. 

In addition the architectural singularity distance $D$ is an upper bound for the distance $d$
of a given robot configuration to its next singularity with respect to the extrinsic metric of \cite[Eq.\ (3)]{b5}; i.e.\ $d\leq D$ has to hold for all configurations of the pentapod. 
As a consequence, we can also come up with a dimensionless distance measure to the closest singularity by computing
$\tfrac{d}{D}\in[0;1]$.

\begin{rmk}\label{rmk:d_D}
In this context, the interesting question arises if configurations with $d=D$ exist? \hfill $\diamond$
\end{rmk}

As the value of $D$ depends on the scaling unit, it cannot be used directly to compare different designs. One can scope this problem by rescaling the given manipulator in a way that 
$max(\rho_{1},\rho_{2})=1$ holds, where $\rho_{1}$ (resp. $\rho_2$) denotes the radius of the smallest sphere (e.g.\ \cite{b8}) enclosing the five base (resp.\ platform) anchor points. {Hence, the scaled version of $D$ can also be seen as a global kinematic performance index}.

Our presented algorithm will not only compute the 
architecture singularity distance but also the closest architectural singularity. This additional information can be exploited to optimize the given design. This can be achieved by utilizing similar ideas as presented in  {\cite{b1}}, where an algorithm is given for optimizing a linear pentapod's configuration in terms of singularity distance.

The distance function $D$ of Eq.\ (\ref{eq:distance0}) can be simplified for most of the classes of architecture singularity, which is discussed next.

\section{Classes of architecture singularity}\label{distances}

\begin{figure*}[]
\begin{center}
    \begin{subfigure}[b]{0.23\textwidth}
        \centering
        \begin{overpic}[width=\linewidth]{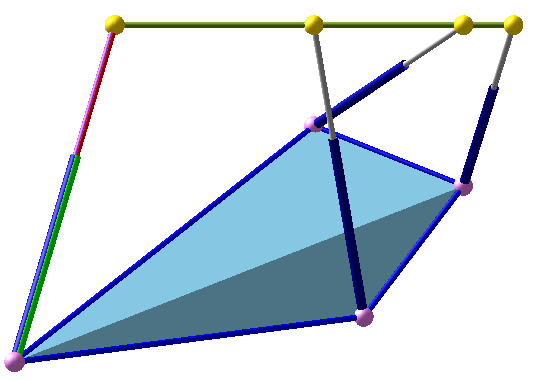}
            \begin{small}
\put(-12,-6){$\mathbf{M}'_{1}=\mathbf{M}'_{2}$}
\put(8,70){$\mathbf{m}'_{1}=\mathbf{m}'_{2}$}
\end{small}
        \end{overpic}
        \vspace{0.2em}
        \caption{Case 0}
    \end{subfigure}
   \hfill
    \begin{subfigure}[b]{0.23\textwidth}
        \centering
        \begin{overpic}[width=\linewidth]{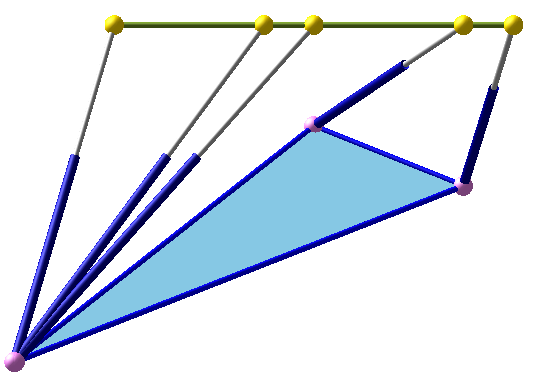}
         \begin{small}
\put(-12,-4){$\mathbf{M'}_{1}=\mathbf{M'}_{2}=\mathbf{M'}_{3}$}
\end{small}  
        \end{overpic}
        \vspace{0.2em}
        \caption{Case 1}
   \end{subfigure}
     \hfill
    \begin{subfigure}[b]{0.23\textwidth}
        \centering
        \begin{overpic}[width=\linewidth]{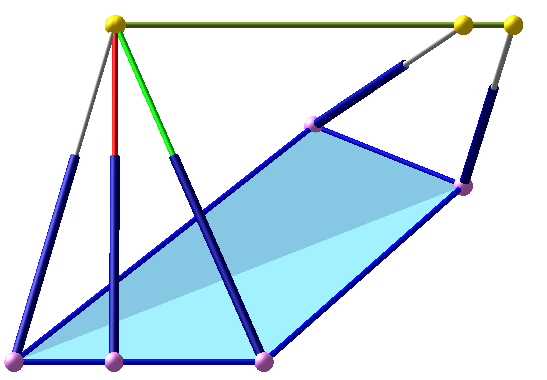}
        \begin{small}
\put(-6,-6){$\mathbf{M}'_{1}$}
\put(14,-6){$\mathbf{M}'_{2}$}
\put(40,-6){$\mathbf{M}'_{3}$}
\put(-10,70){$\mathbf{m}'_{1}=\mathbf{m}'_{2}=\mathbf{m}'_{3}$}
\end{small}   
    \end{overpic}
    \vspace{0.2em}
        \caption{Case 2}
   \end{subfigure}
  \hfill
   \begin{subfigure}[b]{0.23\textwidth}
        \centering
        \begin{overpic}[width=\linewidth]{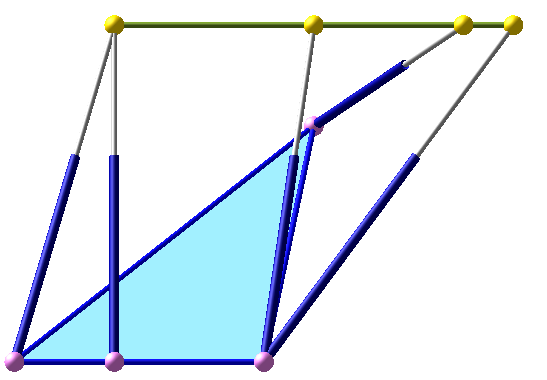}
        \begin{small}
\put(5,70){$\mathbf{m}'_{1}=\mathbf{m}'_{2}$}
\put(40,-6){$\mathbf{M}'_{3}$=$\mathbf{M}'_{4}$}
\put(-8,-6){$\mathbf{M}'_{1}$}
\put(12,-6){$\mathbf{M}'_{2}$}
\end{small}
    \end{overpic}
    \vspace{0.2em}
        \caption{Case 3a}
   \end{subfigure}
\hfill
\vspace{1.5em}
 \begin{subfigure}[b]{0.23\textwidth}
        \centering
        \begin{overpic}[width=\linewidth]{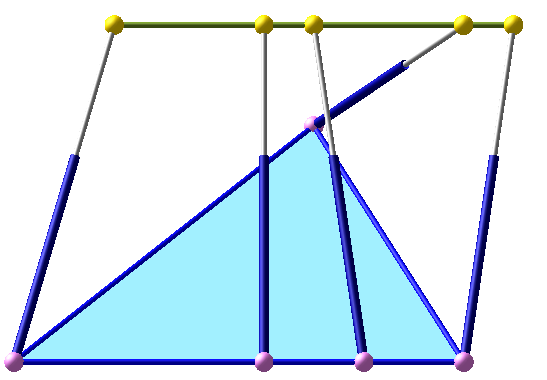}
         \begin{small}
\put(-12,-6){$\mathbf{M}'_{1}$}
\put(40,-6){$\mathbf{M}'_{2}$}
\put(65,-6){$\mathbf{M}'_{3}$}
\put(80,-6){$\mathbf{M}'_{4}$}
\put(15,70){$\mathbf{m}'_{1}$}
\put(40,70){$\mathbf{m}'_{2}$}
\put(55,70){$\mathbf{m}'_{3}$}
\put(90,70){$\mathbf{m}'_{4}$}
\end{small}  
    \end{overpic}
    \vspace{0.2em}
        \caption{Case 3b}
   \end{subfigure}
\hfill
\begin{subfigure}[b]{0.23\textwidth}
        \centering
        \begin{overpic}[width=\linewidth]{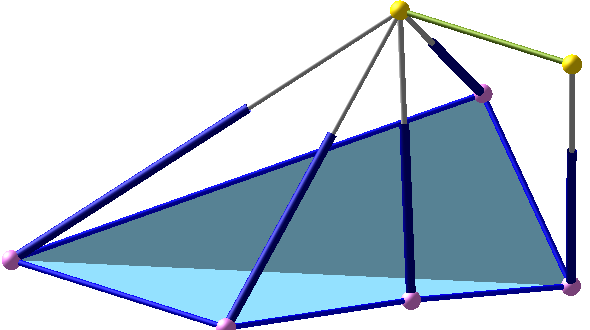}
        \begin{small}
\put(24,58){$\mathbf{m}'_{1}=\mathbf{m}'_{2}=\mathbf{m}'_{3}=\mathbf{m}'_{4}$}
\end{small}
    \end{overpic}
     \vspace{0.2em}
        \caption{Case 4}
   \end{subfigure}
\hfill
\begin{subfigure}[b]{0.23\textwidth}
        \centering
        \begin{overpic}[width=\linewidth]{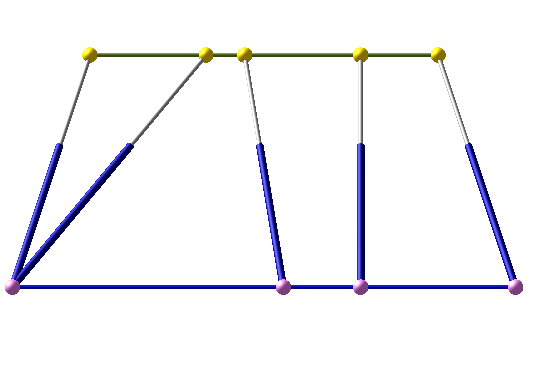}
         \begin{small}
  \put(0,7)
{$\mathbf{M}'_{1}=\mathbf{M}'_{2}$}
\put(50,7){$\mathbf{M}'_{3}$}
\put(68,7){$\mathbf{M}'_{4}$}
\put(90,7){$\mathbf{M}'_{5}$}
\end{small}  
    \end{overpic}
     \vspace{0.2em}
        \caption{Case 5a}
   \end{subfigure}
\hfill
\begin{subfigure}[b]{0.23\textwidth}
        \centering
        \begin{overpic}[width=\linewidth]{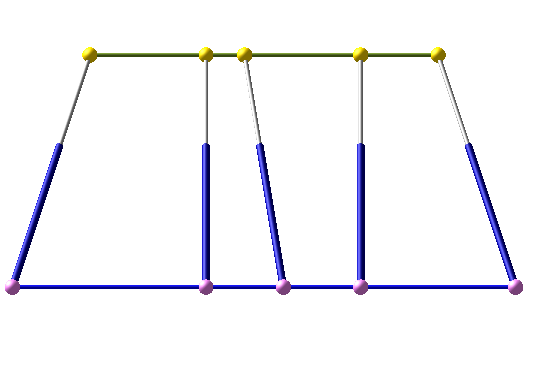}
          \begin{small}
\put(-5,8){$\mathbf{M}'_{1}$}
\put(30,8){$\mathbf{M}'_{2}$}
\put(50,8){$\mathbf{M}'_{3}$}
\put(66,8){$\mathbf{M}'_{4}$}
\put(90,8){$\mathbf{M}'_{5}$}
\end{small}  
    \end{overpic}
     \vspace{0.2em}
        \caption{Case 5b}
   \end{subfigure}
\hfill
\vspace{1em}
 \begin{subfigure}[b]{0.23\textwidth}
        \centering
        \begin{overpic}[width=\linewidth]{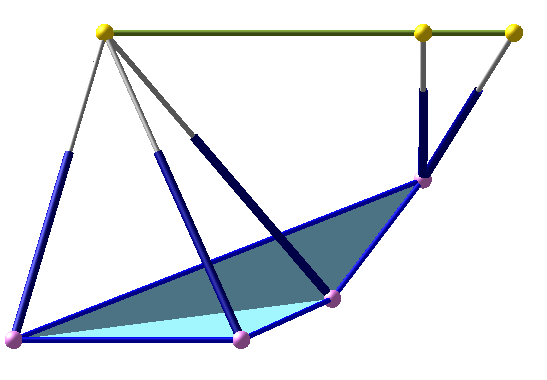}
        \begin{small}
\put(0,70){$\mathbf{m}'_{1}=\mathbf{m}'_{2}=\mathbf{m}_{3}$}
\put(80,31){$\mathbf{M}'_{4}=\mathbf{M}'_{5}$}
\end{small} 
    \end{overpic}
     \vspace{0.1em}
        \caption{Case 6}
   \end{subfigure}
\hfill
\begin{subfigure}[b]{0.23\textwidth}
        \centering
        \begin{overpic}[width=\linewidth]{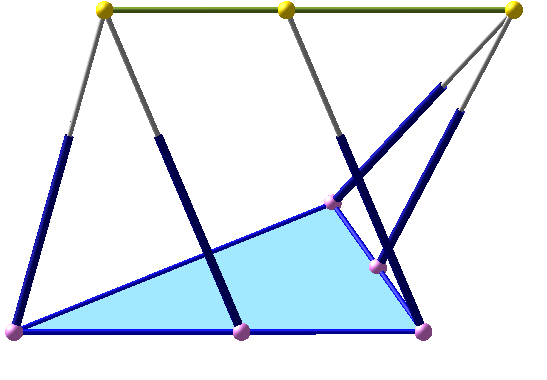}
         \begin{small}
\put(6,72){$\mathbf{m}'_{1}=\mathbf{m}'_{2}$}
\put(50,72){$\mathbf{m}'_{3}$}
\put(76,72){$\mathbf{m}'_{4}=\mathbf{m}'_{5}$}
\put(-6,0){$\mathbf{M}'_{1}$}
\put(40,0){$\mathbf{M}'_{2}$}
\put(80,0){$\mathbf{M}'_{3}$}
\put(75,20){$\mathbf{M}'_{4}$}
\put(49,36){$\mathbf{M}'_{5}$}
\end{small} 
    \end{overpic}
     \vspace{0.1em}
        \caption{Case 7}
   \end{subfigure}
\hfill
\begin{subfigure}[b]{0.23\textwidth}
        \centering
        \begin{overpic}[width=\linewidth]{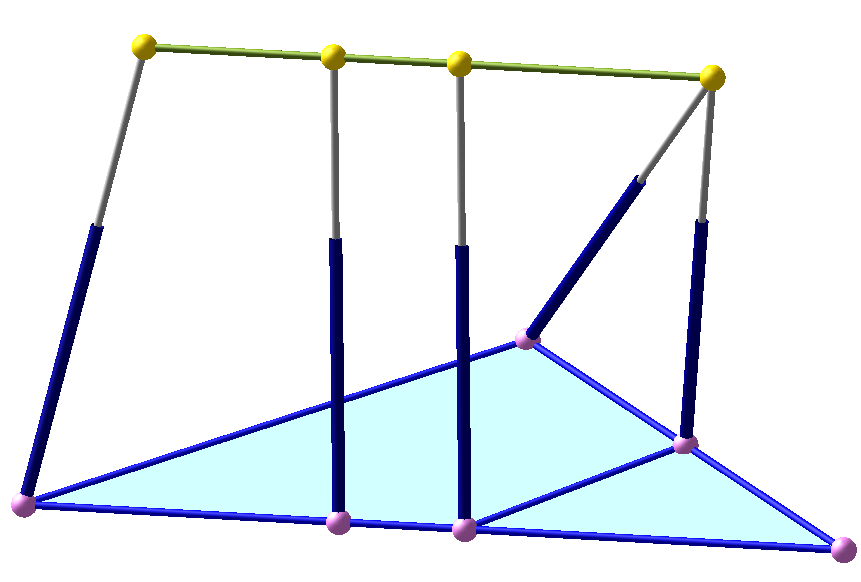}
        \begin{small}
\put(68,65){$\mathbf{m}'_{4}=\mathbf{m}'_{5}$}
\put(-5,0){$\mathbf{M}'_{1}$}
\put(30,-3){$\mathbf{M}'_{2}$}
\put(50,-3){$\mathbf{M}'_{3}$}
\put(84,15){$\mathbf{M}'_{4}$}
\put(66,25.5){$\mathbf{M}'_{5}$}
\put(94,-4){$\mathbf{M}$}
\end{small} 
    \end{overpic}
     \vspace{0.1em}
        \caption{Case 8}
   \end{subfigure}
\hfill
\begin{subfigure}[b]{0.23\textwidth}
        \centering
        \begin{overpic}[width=\linewidth]{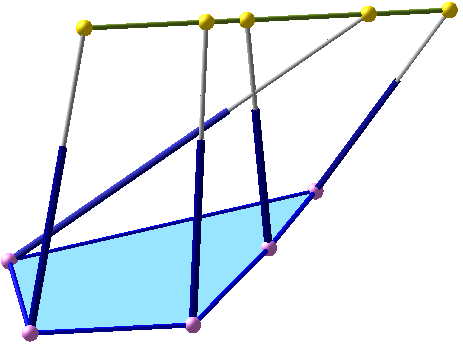}
         \begin{small}
\put(-5,-5){$\mathbf{M}'_{1}$}
\put(42,-4){$\mathbf{M}'_{2}$}
\put(60,15){$\mathbf{M}'_{3}$}
\put(70,30){$\mathbf{M}'_{5}$}
\put(-5,25){$\mathbf{M}'_{4}$}
\put(13,75){$\mathbf{m}'_{1}$}
\put(38,75){$\mathbf{m}'_{2}$}
\put(50,75){$\mathbf{m}'_{3}$}
\put(72,77){$\mathbf{m}'_{4}$}
\put(90,77){$\mathbf{m}'_{5}$}
\end{small}
    \end{overpic}
     \vspace{0.1em}
        \caption{Case 9}
   \end{subfigure}
\end{center}
\caption{Illustration of architectural singular designs of linear pentapods.} 
\label{designs}
\end{figure*}
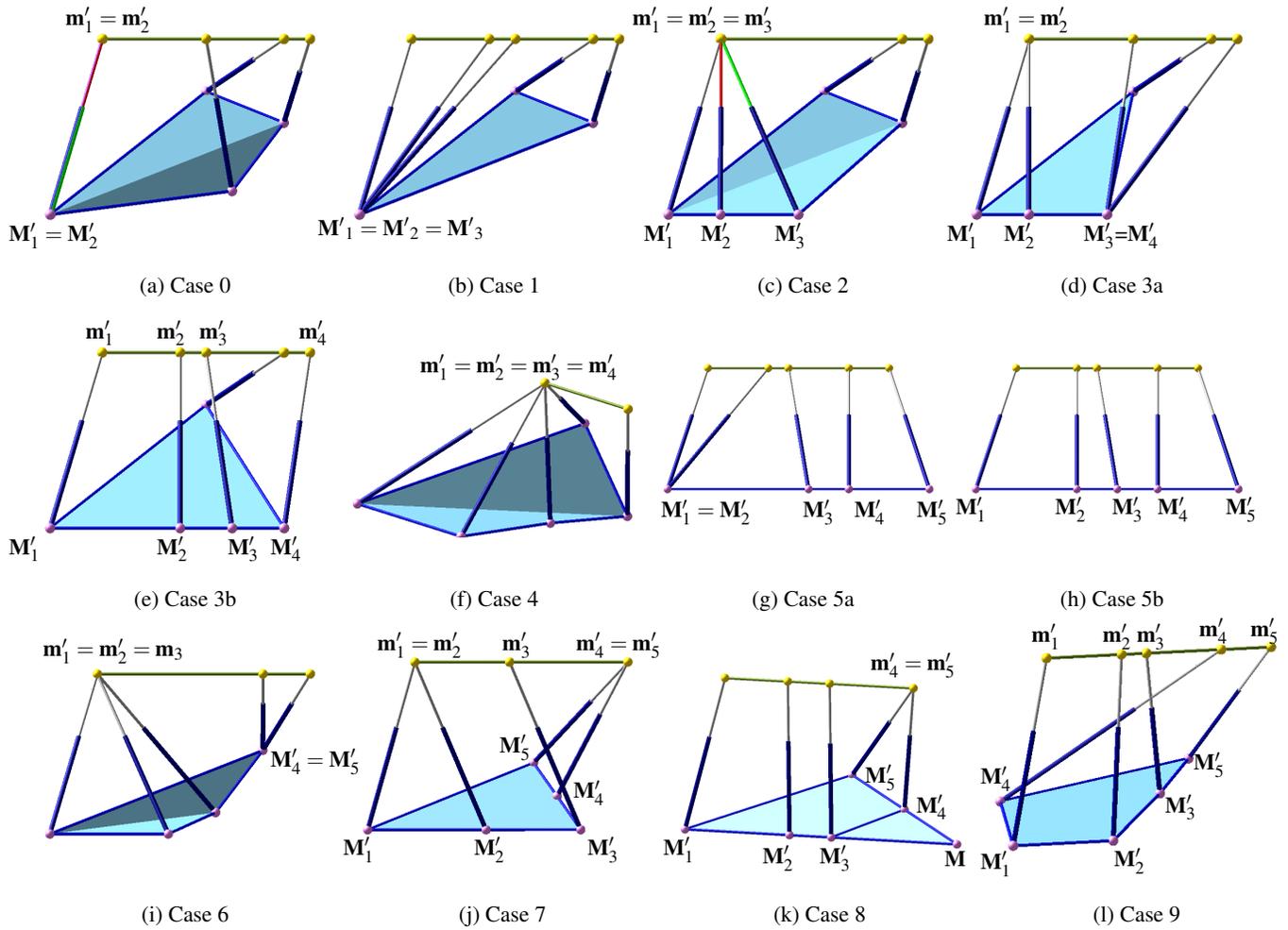

If a linear pentapod is architecturally singular then it has to be one of the nine designs (1-9) listed in \cite{b4} up to a possible necessary renumbering of anchor points. Fig.\ \ref{designs} illustrates all nine design cases of architecturally singular pentapods.

We add the trivial case of two coinciding legs (i.e.\ the pentapod degenerates to a quadropod) as case 0 to this list, which is now analyzed point by point regarding their computational treatment. {Note that
for the discussion of any of these cases, we can always exclude the designs of all previous cases.}

\paragraph*{Case 0}  $\mathbf{M}^{'}_{1}=\mathbf{M}^{'}_{2}$ and  $\mathbf{m}^{'}_{1}=\mathbf{m}^{'}_{2}$: 
As there are no conditions on the remaining points $\mathbf{M}^{'}_{j}$ and $\mathbf{m}^{'}_{j}$ for $j=3,4,5$ the distance function simplifies to 
\begin{equation*}
D^2_{0}= {\frac{1}{10}\sum_{i=1}^{2}\left[||{\mathbf{M}^{'}_{i}}-\mathbf{M}_{i}||^{2}+||\mathbf{m}^{'}_{i}-\mathbf{m}_{i}||^{2}\right]},
\end{equation*} 
thus $D^2_{0}$ only depends on the four unknowns $x'_{1}, y'_{1}, z'_{1}, r'_{1}$.

\paragraph*{Case 1} $\mathbf{M}^{'}_{1} = \mathbf{M}^{'}_{2}=\mathbf{M}^{'}_{3}$:
In this case, the distance function reduces to 
\begin{equation*}
D^2_{1} = {\frac{1}{10}\sum_{i=1}^{3}||{\mathbf{M}^{'}_{i}}-\mathbf{M}_{i}||^{2}},
\end{equation*} 
thus $D^2_{1}$ only depends on the three  unknowns $x'_{1}, y'_{1}, z'_{1}$.

\paragraph*{Case 2} $\mathbf{m}^{'}_{1}=\mathbf{m}^{'}_{2}=\mathbf{m}^{'}_{3} $  and $\mathbf{M}^{'}_{1}, \mathbf{M}^{'}_{2}, \mathbf{M}^{'}_{3}$ are collinear: 
As $\mathbf{M}^{'}_{1}=\mathbf{M}^{'}_{2}$ implies case 0, we can assume without loss of generality (w.l.o.g.) that $\mathbf{M}^{'}_{1}\neq\mathbf{M}^{'}_{2}$ holds. 
Therefore, $\mathbf{M}^{'}_{3}$ can be parametrized as 
\begin{equation}
\mathbf{M}^{'}_{3}=\mathbf{M}^{'}_{1}+\Lambda(\mathbf{M}^{'}_{2}-\mathbf{M}^{'}_{1}),
\label{p1}
\end{equation}
and the distance function is reduced to
\begin{equation*}
D^2_{2} = \frac{1}{10}\sum_{i=1}^{3}\left[||{\mathbf{M}^{'}_{i}}-\mathbf{M}_{i}||^{2}+||\mathbf{m}^{'}_{i}-\mathbf{m}_{i}||^{2}\right], 
\end{equation*}
which only depends on the eight unknowns $r'_{1},\Lambda,x'_{i}, y'_{i}, z'_{i}$ for $i=1,2$.

\paragraph*{Case 3} $\mathbf{M}^{'}_{1}, \ldots ,\mathbf{M}^{'}_{4}$ are collinear and the following cross-ratio condition holds: 
\begin{equation}
CR(\mathbf{m}^{'}_{1}, \mathbf{m}^{'}_{2}, \mathbf{m}^{'}_{3}, \mathbf{m}^{'}_{4})= CR(\mathbf{M}^{'}_{1}, \mathbf{M}^{'}_{2}, \mathbf{M}^{'}_{3}, \mathbf{M}^{'}_{4}).
\label{cross12}
\end{equation}
From the computational point of view, we have to divide this case into the following two subcases:
\begin{enumerate}[(a)]

\item
$\mathbf{m}^{'}_{1}=\mathbf{m}^{'}_{2}$: Under this assumption the cross-ratio condition of Eq.\ (\ref{cross12}) implies  $\mathbf{M}^{'}_{3}=\mathbf{M}^{'}_{4}$. 
As $\mathbf{M}^{'}_{1}=\mathbf{M}^{'}_{2}$ implicate case 0, we can assume w.l.o.g.\ that $\mathbf{M}^{'}_{1}\neq\mathbf{M}^{'}_{2}$ holds which allows to 
parametrize $\mathbf{M}^{'}_{3}=\mathbf{M}^{'}_{4}$ in analogy to Eq.\ (\ref{p1}). Now the distance function is reduced to 
\begin{equation*}
D^{2}_{3a}={\frac{1}{10}\left[\sum_{i=1}^{2}||{\mathbf{m}^{'}
_{i}}-\mathbf{m}_{i}||^{2}+\sum_{i=1}^{4}||{\mathbf{M}^{'}_{i}}-\mathbf{M}_{i}||^{2}\right]} ,
\end{equation*}
which depends on the same unknowns as $D^2_{2}$ of case 2.

\begin{rmk}
The case $\mathbf{M}^{'}_{1}=\mathbf{M}^{'}_{2}$ leads to the analogous case as already discussed in case 3(a).  One just has to swap the indices 
$1$ and $3$ as well as $2$ and $4$. \hfill $\diamond$
\end{rmk}

\item
By assuming $\mathbf{M}^{'}_{1}\neq\mathbf{M}^{'}_{2}$ and $\mathbf{m}^{'}_{1}\neq\mathbf{m}^{'}_{2}$,  the points $\mathbf{M}^{'}_{4},\mathbf{m}^{'}_{3}$ 
and $\mathbf{m}^{'}_{4}$ can be parametrized as  
\begin{align}
\mathbf{M}^{'}_{4}&=\mathbf{M}^{'}_{1}+ \Delta(\mathbf{M}^{'}_{2}-\mathbf{M}^{'}_{1}), \label{p3} \\
\mathbf{m}^{'}_{3}&=\mathbf{m}^{'}_{1}+\lambda(\mathbf{m}^{'}_{2}-\mathbf{m}^{'}_{1}),  \quad
\mathbf{m}^{'}_{4}=\mathbf{m}^{'}_{1}+\delta (\mathbf{m}^{'}_{2}-\mathbf{m}^{'}_{1}),  \label{p45}
 \end{align}
and $\mathbf{M}^{'}_{3}$ as given by Eq.\ (\ref{p1}). With respect to this parametrization the side-condition $S=0$ stated in Eq.\ (\ref{cross12})
reads as:
\begin{equation}
    S=\lambda\delta(\Lambda-\Delta)+\lambda(\Delta-\Lambda\Delta)+\delta(\Lambda\Delta-\Lambda)
    \label{side1}
\end{equation}
and the distance function is reduced to
\begin{equation*}
D^{2}_{3b} = {\frac{1}{10}\sum_{i=1}^{4}\left[||{\mathbf{M}^{'}_{i}}-\mathbf{M}_{i}||^{2}+||{\mathbf{m}^{'}_{i}}-\mathbf{m}_{i}||^{2}\right]}.
\end{equation*}
Due to the side condition, the minimization problem has to be solved by the Lagrangian approach 
\begin{equation}
L_{3b}=D_{3b}^{2}+\mu S.
\label{c1}
\end{equation}
Note that $L_{3b}$ depends on the thirteen unknowns $\mu, \Lambda, \Delta, \lambda, \delta, x'_{i}, y'_{i}, z'_{i}, r'_{i}$ for $i=1,2$.

It is well known (e.g.\ \cite{b1}) 
that singular points of the constrained variety $S=0$ are excluded from the Lagrangian approach given in Eq.\ (\ref{c1}). Therefore, we have to identify these points, which can be done by computing the partial derivatives of $S$ with respect to the 4 unknowns $\lambda, \Delta, \Lambda, \delta$. It can easily be seen that all real solutions\footnote{There are also conjugate solutions which do not cause any coincidence of legs.} of the resulting system 
are causing the coincidence of legs; either three legs coincide or two times two legs coincide. But these are special cases of case 0, thus we do not have to consider them.
\end{enumerate}
\paragraph*{Case 4} $\mathbf{m}^{'}_{1}=\mathbf{m}^{'}_{2}=\mathbf{m}^{'}_{3}=\mathbf{m}^{'}_{4}$: 
The distance function simplifies to
\begin{equation*}
D^2_{4} = {\frac{1}{10}\sum_{i=1}^{4}||{\mathbf{m}^{'}_{i}}-\mathbf{m}_{i}||^{2}},
\end{equation*}
thus $D^2_{4}$ only depends on $r'_{1}$. 
\paragraph*{Case 5} $\mathbf{M}^{'}_{1}, \ldots ,\mathbf{M}^{'}_{5}$ are collinear: 
We have to distinguish the following two subcases:

\begin{enumerate}[(a)]
\item
$\mathbf{M}^{'}_{1}=\mathbf{M}^{'}_{2}$: Now we can assume w.l.o.g.\ that $\mathbf{M}^{'}_{1} \neq \mathbf{M}^{'}_{3}$ holds, as otherwise we would end up with 
case 1, which allows to make the following parametrization
\begin{equation*}
\mathbf{M}^{'}_{4}=\mathbf{M}^{'}_{1}+\Gamma(\mathbf{M}^{'}_{3}-\mathbf{M}^{'}_{1}), \quad
\mathbf{M}^{'}_{5}=\mathbf{M}^{'}_{1}+\Phi(\mathbf{M}^{'}_{3}-\mathbf{M}^{'}_{1}). 
\end{equation*}
The distance function reduces to
\begin{equation}
D^2_{5a} = {\frac{1}{10}\sum_{i=1}^{5}||{\mathbf{M}^{'}_{i}}-\mathbf{M}_{i}||^{2}}, 
\label{eq:distance7}
\end{equation}
which depends on the eight unknowns $\Gamma, \Phi, x'_{i}, y'_{i}, z'_{i}$ for $i=1,3$.
\item
Assuming $\mathbf{M}^{'}_{1}\neq\mathbf{M}^{'}_{2}$ the points  $\mathbf{M}^{'}_{3}$ and  $\mathbf{M}^{'}_{4}$ can be parametrized as in 
Eqs.\ (\ref{p1}) and (\ref{p3}) then $\mathbf{M}^{'}_{5}$ is given as
\begin{equation*}
\mathbf{M}^{'}_{5}=\mathbf{M}^{'}_{1}+\Phi(\mathbf{M}^{'}_{2}-\mathbf{M}^{'}_{1}).
\end{equation*}
The distance function $D_{5b}$ remains the same as in Eq.\ (\ref{eq:distance7}), {but now it depends on nine unknowns, which are  $\Lambda, \Delta, \Phi,x'_{i}, y'_{i}, z'_{i}$ for $i=1,2$.}
\end{enumerate}

\paragraph*{Case 6} $\mathbf{m}^{'}_{1}=\mathbf{m}^{'}_{2}=\mathbf{m}^{'}_{3}$ and $\mathbf{M}^{'}_{4} = \mathbf{M}^{'}_{5}$. In this case, the distance function reduces to
\begin{equation*}
D^{2}_{6} = {\frac{1}{10}\left[\sum_{i=4}^{5}||{\mathbf{M}^{'}_{i}}-\mathbf{M}_{i}||^{2}+\sum_{i=1}^{3}||{\mathbf{m}^{'}_{i}}-\mathbf{m}_{i}||^{2}\right]} 
\end{equation*}
thus $D^2_{6}$ depends on the four unknowns $x'_{4}, y'_{4}, z'_{4},r'_{1}$.
\paragraph*{Case 7} $\mathbf{m}^{'}_{1}=\mathbf{m}^{'}_{2}$ and $\mathbf{m}^{'}_{4}=\mathbf{m}^{'}_{5}$ holds and
$\mathbf{M}^{'}_{1}, \mathbf{M}^{'}_{2}, \mathbf{M}^{'}_{3}$ as well as $\mathbf{M}^{'}_{3},\mathbf{M}^{'}_{4},\mathbf{M}^{'}_{5}$ are collinear: 
W.l.o.g.\ one can assume that $\mathbf{M}^{'}_{3}\neq\mathbf{M}^{'}_{4}$ holds as otherwise we would end up with a special case of case 3b ($\mathbf{m}^{'}_{1}=\mathbf{m}^{'}_{2}$ and $\mathbf{M}^{'}_{3}=\mathbf{M}^{'}_{4}$). For the same reasoning (just for another indexing) one can also assume w.l.o.g.\ that $\mathbf{M}^{'}_{2}\neq\mathbf{M}^{'}_{3}$ holds, which allows 
the parametrization
\begin{equation*}
\mathbf{M}^{'}_{1}=\mathbf{M}^{'}_{3}+{\Gamma}(\mathbf{M}^{'}_{2}-\mathbf{M}^{'}_{3}), \quad
\mathbf{M}^{'}_{5}=\mathbf{M}^{'}_{3}+\Phi(\mathbf{M}^{'}_{4}-\mathbf{M}^{'}_{3}). 
\end{equation*}
The distance function simplifies to
\begin{equation*}
D^{2}_{7} =\frac{1}{10}\left[\sum_{i=1}^{5}||{\mathbf{M}^{'}_{i}}-\mathbf{M}_{i}||^{2}+
\sum_{i=1,2,4,5}||{\mathbf{m}^{'}_{i}}-\mathbf{m}_{i}||^{2} \right],
\end{equation*}
{
which depends on the thirteen unknown $\Gamma, \Phi,r'_{1}, r'_{2},x'_{i}, y'_{i}, z'_{i}$ for $i=2,3,4$.}

\paragraph*{Case 8} $\mathbf{m}^{'}_{4}=\mathbf{m}^{'}_{5}$ and $\mathbf{M}^{'}_{1}, \ldots ,\mathbf{M}^{'}_{5}$ are coplanar where  $\mathbf{M}^{'}_{1}, \mathbf{M}^{'}_{2}, \mathbf{M}^{'}_{3}$ are collinear and the following cross-ratio condition holds:
\begin{equation}
CR(\mathbf{m}^{'}_{1}, \mathbf{m}^{'}_{2}, \mathbf{m}^{'}_{3}, \mathbf{m}^{'}_{4})= CR(\mathbf{M}^{'}_{1}, \mathbf{M}^{'}_{2}, \mathbf{M}^{'}_{3}, \mathbf{M}^{'})
\label{cross123}
\end{equation}
with $\mathbf{M}^{'}$ denoting the intersection point of the line spanned by $\mathbf{M}^{'}_{4}, \mathbf{M}^{'}_{5}$ and the carrier line of  
$\mathbf{M}^{'}_{1}, \mathbf{M}^{'}_{2}, \mathbf{M}^{'}_{3}$. 
W.l.o.g.\ we can assume $\mathbf{M}^{'}_{4}\neq \mathbf{M}^{'}_{5}$ (as otherwise we get case 0), which allows to parametrize $\mathbf{M}^{'}$ by 
\begin{equation*}
\mathbf{M}^{'}=\mathbf{M}^{'}_{4}+\Gamma(\mathbf{M}^{'}_{5}-\mathbf{M}^{'}_{4}).
\end{equation*}
Moreover, one can assume $\mathbf{m}^{'}_{1}\neq\mathbf{m}^{'}_{4}$ as otherwise  the cross-ratio condition of Eq.\ (\ref{cross123}) implies 
$\mathbf{M}^{'}_{2}=\mathbf{M}^{'}_{3}$ which yields the already discussed case 6 up to an exchange of indices.
Therefore, one can parametrize the points $\mathbf{m}^{'}_{2}$ and $\mathbf{m}^{'}_{3}$ by 
\begin{equation*}
\mathbf{m}^{'}_{2}=\mathbf{m}^{'}_{4}+\lambda(\mathbf{m}^{'}_{1}-\mathbf{m}^{'}_{4}),\quad
\mathbf{m}^{'}_{3}=\mathbf{m}^{'}_{4}+\delta(\mathbf{m}^{'}_{1}-\mathbf{m}^{'}_{4}).
\end{equation*}
W.l.o.g.\ we can also assume $\mathbf{M}^{'}_{1}\neq\mathbf{M}^{'}$ as otherwise the condition given in Eq.\ (\ref{cross123}) 
implies $\mathbf{m}^{'}_{2}=\mathbf{m}^{'}_{3}$ which yields case 7 up to an exchange of indices. Therefore we can set:
\begin{equation*}
\mathbf{M}^{'}_{2}=\mathbf{M}^{'}+\Lambda(\mathbf{M}^{'}_{1}-\mathbf{M^{'}}), \quad 
\mathbf{M}^{'}_{3}=\mathbf{M}^{'}+\Delta(\mathbf{M}^{'}_{1}-\mathbf{M}^{'}). 
\end{equation*}
 With respect to this parametrization, the side-condition  stated in Eq.\ (\ref{cross123})
reads as $S=0$ with $S$ of Eq.\ (\ref{side1}).
In this case the distance function given in Eq.\ (\ref{eq:distance0}) cannot be simplified; i.e. $D_8:=D$, thus the Lagrangian function of the 
optimization problem equals 
\begin{equation*}
L_{8}=D_8^{2}+\mu S.
\end{equation*}

Note that $L_{8}$ depends on the seventeen unknowns $\mu,\Gamma, \Lambda, \Delta, \lambda, \delta,r'_{1}, r'_{4},x'_{i}, y'_{i}, z'_{i}$ for $i=1,4,5$.

Similarly to case 3b, also in this case the real singular points of $S$ imply the coincidence of two legs already covered by case 0.

\paragraph*{Case 9}  $\mathbf{m}^{'}_{1}, \ldots ,\mathbf{m}^{'}_{5}$ are pairwise distinct and $\mathbf{M}^{'}_{1}, \ldots ,\mathbf{M}^{'}_{5}$ are coplanar where no three of them are collinear and 
there is a projective correspondence $\sigma$ between the platform points and the base points:
Therefore, one can chose the points $\mathbf{M}^{'}_{1}, \mathbf{M}^{'}_{2}, \mathbf{M}^{'}_{3} $ and $\mathbf{m}^{'}_{1}, \mathbf{m}^{'}_{2}$ as free variables and 
parameterize the remaining anchor points in their dependence by:
\begin{align*}
\mathbf{M}^{'}_{4}&=\mathbf{M}^{'}_{1}+ \Psi_{1}(\mathbf{M}^{'}_{2}-\mathbf{M}^{'}_{1})+ \Upsilon_{1} (\mathbf{M}^{'}_{3}-\mathbf{M}^{'}_{1}), \\ 
\mathbf{M}^{'}_{5}&=\mathbf{M}^{'}_{1}+\Psi_{2}(\mathbf{M}^{'}_{2}-\mathbf{M}^{'}_{1})+ \Upsilon_{2} (\mathbf{M}^{'}_{3}-\mathbf{M}^{'}_{1}), \\
\mathbf{m}^{'}_{5}&=\mathbf{m}^{'}_{1}+\gamma(\mathbf{m}^{'}_{2}-\mathbf{m}^{'}_{1}),
\end{align*}
and $\mathbf{m}^{'}_{3}$ and  $\mathbf{m}^{'}_{4}$ are given by  Eq.\ (\ref{p45}).
As it is well-known \cite{b4} that the property of being architecture singular remains invariant under separate projective transformations of the platform and the base,  
one can assign special coordinate frames to the platform and the base such that the correspondence $\sigma$  
can be expressed by the two algebraic conditions $S_1=0$ and $S_2=0$ (cf.\  \cite[Eq.\ (10)]{b3} as well as \cite{b7}) with
\begin{equation*}
{S}_1= 
\begin{small}
 \begin{vmatrix}
1 & 0 & 0 & 0 & 0 \\ 
1 & 1 & 0 & 1 & 1  \\ 
1 & 0 & 1 & 0 & \lambda \\ 
1 & \Psi_{1} &\Upsilon_{1} & \Psi_{1}\delta & \delta\\
1 & \Psi_{2} &\Upsilon_{2} & \Psi_{2}\gamma & \gamma  
\end{vmatrix}
\end{small}, \quad
{S}_2= 
\begin{small}
\begin{vmatrix}
1 & 0 & 0 & 0 & 0 \\ 
1 & 1 & 0 & 1 & 0  \\ 
1 & 0 & 1 & 0 & \lambda \\ 
1 & \Psi_{1} &\Upsilon_{1} & \Psi_{1}\delta & \Upsilon_{1} \delta\\
1 & \Psi_{2} &\Upsilon_{2} & \Psi_{2}\gamma & \Upsilon_{2} \gamma   
\end{vmatrix}
\end{small}. 
\end{equation*}
Note that case 9 is the most general one, thus the distance function given in Eq.\ (\ref{eq:distance0}) cannot be reduced; i.e. $D_9:=D$.  
Due to the existence of two side conditions the minimization problem has to be solved by the Lagrangian approach 
\begin{equation}
L_{9}=D_9^{2}+\mu_1 S_{1}+\mu_2 S_{2}.
\label{c2}
\end{equation}

Note that $L_{9}$ depends on the twenty unknowns  $\lambda, \delta, \gamma,r_j,\mu_j,\Upsilon_{j}, \Psi_{j},x'_{i}, y'_{i}, z'_{i}$ for $i=1,2,3$, and $j=1,2$. 

Also in this case the singular points of the constraint varieties $S_{1}=S_{2}=0$ are excluded from the Lagrangian approach given in Eq.~(\ref{c2}).  
These points can be identified algebraically by the system of equations obtained by computing the partial derivatives of
{
\begin{equation}
\mu_{1}S_{1}+\mu_{2}S_{2} 
\label{main3}
\end{equation}}
with respect to the 9 unknowns, $\mu_j$, $\Psi_{j}$, $\Upsilon_{j}$, $\gamma$, $\delta$, $\lambda$.
They form the following ideal:
\begin{equation}
\langle g_{1}, \dots, g_{9}\rangle \subseteq \mathbb{K}[\Psi_{1}, \Upsilon_{1},  \Psi_{2}, \Upsilon_{2},\gamma, \delta, \lambda, \mu_{1},\mu_{2}]. 
\label{idealnew}
\end{equation}

Solving  Eq.~(\ref{idealnew}) using Gr\"obner basis method implemented in \texttt{Maple}, results in $62$ solution sets. Investigating these $62$ solution sets reveals that each of them either has three collinear base points (including the case where two base points coincide) or two coinciding platform points. Thus none of the solutions are valid, as they conflict with the assumption of case 9. {The corresponding \texttt{Maple} file can be downloaded from \cite{b18}.}

\begin{rmk}\label{rmk:planar}
The total number of unknowns for all the mentioned optimization problems is summarized in the $5^{th}$ column of Table~\ref{data}.
In many real-world scenarios, the five base anchor points of a linear pentapod are coplanar. 
This reduces the number of unknowns in the optimization problems as pointed out in  Table~\ref{data}. \hfill $\diamond$  
\end{rmk}

\subsection {Geometric characterization of the minimizers}

Based on the given formulation of the minimization problems for the individual cases, one can give immediately the following geometric characterizations of the minimizers:

\begin{enumerate}[$\bullet$]
    \item 
    {Case 0 : $\mathbf{M}^{'}_{1,2}$  is the mid-point of $\mathbf{M}_{1}$ and $\mathbf{M}_{2}$ and $\mathbf{m}^{'}_{1,2}$ is the mid-point of $\mathbf{m}_{1}$ and $\mathbf{m}_{2}$.}
    \item 
    {Case 1 :
    $\mathbf{M}^{'}_{1,2,3}$ is the centroid of the three base anchor points $\mathbf{M}_{1},\mathbf{M}_{2},\mathbf{M}_{3}$.}
    \item 
    {Case 2:  $\mathbf{m}^{'}_{1,2,3}$ is the centroid of $\mathbf{m}_{1},\mathbf{m}_{2},\mathbf{m}_{3}$. $\mathbf{M}'_{1},\mathbf{M}'_{2},\mathbf{M}'_{3}$ are the pedal points on the line of regression of the points $\mathbf{M}_{1},\mathbf{M}_{2},\mathbf{M}_{3}$.}
    \item 
   {Case 4: $\mathbf{m}^{'}_{1,2,3,4}$ is the centroid of $\mathbf{m}_{1},\mathbf{m}_{2},\mathbf{m}_{3}$, $\mathbf{m}_{4}$.}
    \item 
   {Case 5b: $\mathbf{M}'_{1},\ldots, \mathbf{M}'_{5}$  are the pedal points on the line of regression of the points  $\mathbf{M}_{1},\ldots, \mathbf{M}_{5}$.}
    
    \item {Case 6: $\mathbf{m}^{'}_{1,2,3}$ is the centroid of $\mathbf{m}_{1},\mathbf{m}_{2},\mathbf{m}_{3}$. $\mathbf{M}^{'}_{4,5}$  is the midpoint of $\mathbf{M}_{4}$ and $\mathbf{M}_{5}$}
\end{enumerate}
The geometric interpretations for the minimizers of cases 3b, 5a, 8, and 9 remain open. For cases 3a and 7 we only know that  $\mathbf{m}^{'}_{i,j}$ is the mid-point of $\mathbf{m}_{i}$ and $\mathbf{m}_{j}$, but we could not figure out a geometric meaning for the base points.

\begin{table*}[h!]
 \caption{Summary of computational data for ab-into phase for planar and non-planar architecture singular design cases.}
\centering
\begin{tabular}{|lll||ll||ll|}
\hline
\multicolumn{3}{|l||}{}                                 & \multicolumn{2}{c||}{Planar}     & \multicolumn{2}{c|}{Non-planar} \\ \hline
\multicolumn{1}{|l||}{case $k$} &
  \multicolumn{1}{l|}{degree of $L_{k}$} &
  $\#$ combinations &
  \multicolumn{1}{l|}{ $\#$ unknowns} &
  \begin{tabular}[c]{@{}l@{}} $\#$ finite sol-\\utions  over C\end{tabular} &
  \multicolumn{1}{l|}{ $\#$ uknowns} &
  \begin{tabular}[c]{@{}l@{}} $\#$ finite sol-\\utions  over C\end{tabular} \\ \hline\hline
\multicolumn{1}{|l||}{0}  & \multicolumn{1}{l|}{2} & 10 & \multicolumn{1}{l|}{3}  & 1     & \multicolumn{1}{l|}{4}  & 1     \\ \hline
\multicolumn{1}{|l||}{1}  & \multicolumn{1}{l|}{2} & 10 & \multicolumn{1}{l|}{2}  & 1     & \multicolumn{1}{l|}{3}  & 1     \\ \hline
\multicolumn{1}{|l||}{2}  & \multicolumn{1}{l|}{4} & 10 & \multicolumn{1}{l|}{6}  & 2     & \multicolumn{1}{l|}{8}  & 2     \\ \hline
\multicolumn{1}{|l||}{3a} & \multicolumn{1}{l|}{4} & 30 & \multicolumn{1}{l|}{6}  & 2     & \multicolumn{1}{l|}{8}  & 2     \\ \hline
\multicolumn{1}{|l||}{3b} & \multicolumn{1}{l|}{4} & 5  & \multicolumn{1}{l|}{11} & 88    & \multicolumn{1}{l|}{13} & 88    \\ \hline
\multicolumn{1}{|l||}{4}  & \multicolumn{1}{l|}{2} & 5  & \multicolumn{1}{l|}{1}  & 1     & \multicolumn{1}{l|}{1}  & 1     \\ \hline
\multicolumn{1}{|l||}{5a} & \multicolumn{1}{l|}{4} & 10 & \multicolumn{1}{l|}{6}  & 2     & \multicolumn{1}{l|}{8}  & 2     \\ \hline
\multicolumn{1}{|l||}{5b} & \multicolumn{1}{l|}{4} & 1  & \multicolumn{1}{l|}{7}  & 2     & \multicolumn{1}{l|}{9}  & 2     \\ \hline
\multicolumn{1}{|l||}{6}  & \multicolumn{1}{l|}{4} & 10 & \multicolumn{1}{l|}{3}  & 1     & \multicolumn{1}{l|}{4}  & 1     \\ \hline
\multicolumn{1}{|l||}{7}  & \multicolumn{1}{l|}{5} & 15 & \multicolumn{1}{l|}{10} & 12    & \multicolumn{1}{l|}{13} & 12    \\ \hline
\multicolumn{1}{|l||}{8}  & \multicolumn{1}{l|}{6} & 10 & \multicolumn{1}{l|}{14} & 332   & \multicolumn{1}{l|}{17} & 350   \\ \hline
\multicolumn{1}{|l||}{9}  & \multicolumn{1}{l|}{5} & 1  & \multicolumn{1}{l|}{17} & 2 435 & \multicolumn{1}{l|}{20} & 2\,756 \\ \hline
\end{tabular}
\label{data}
\end{table*}

\section{Computational procedure}\label{sec:procedure1}
\begin{figure*}[htbp!]
\centering
 \begin{overpic}[width=150mm]
 {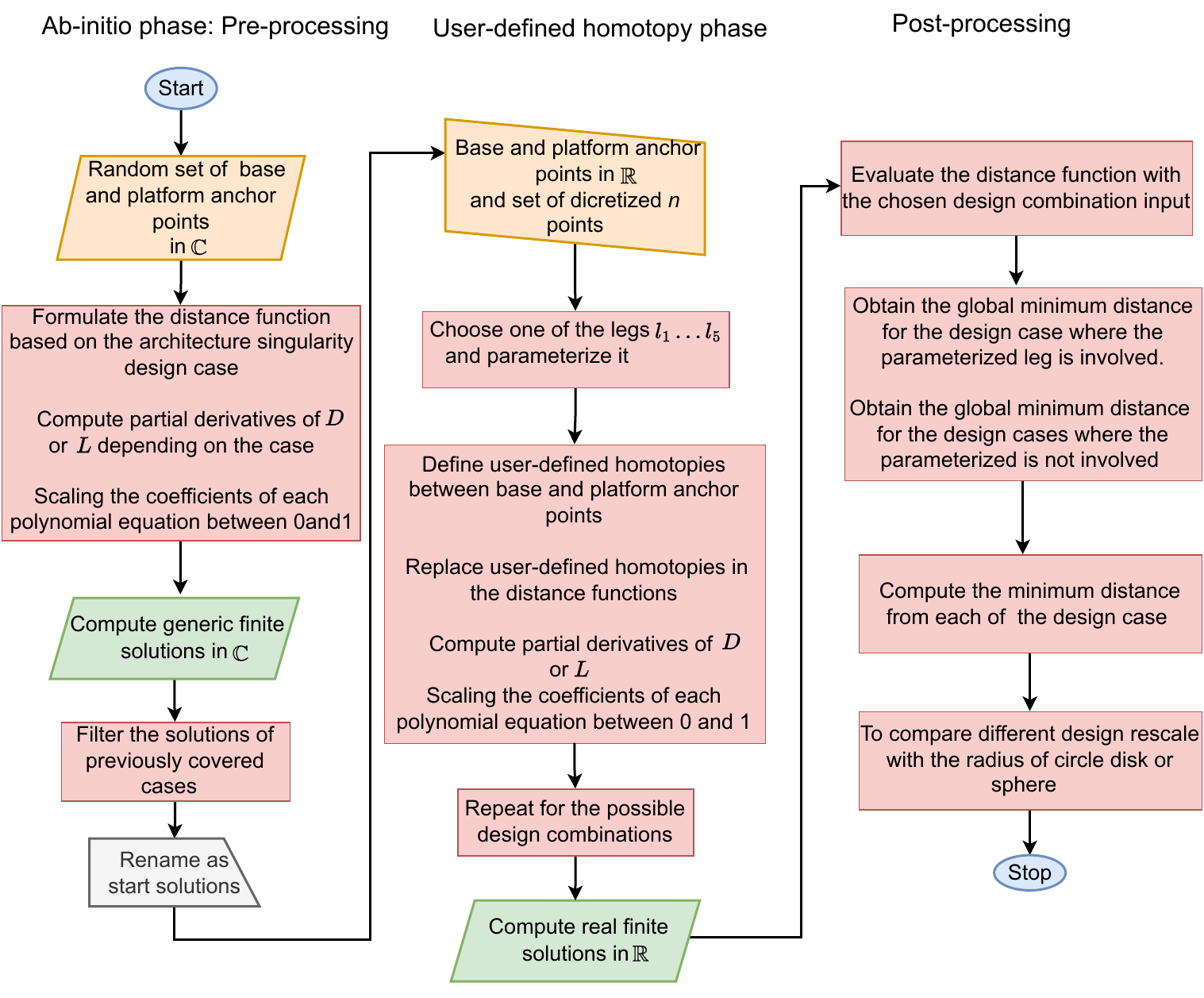}
\end{overpic}
\caption{Homotopy continuation-based computational pipeline for finding the closest architecture singular design.}
\label{fig:flowchart1}
\end{figure*}

In this section, we introduce a computational pipeline, as illustrated in Fig.\ \ref{fig:flowchart1}, that outlines the operational flowchart for finding the closest architectural singularity for a given design based on the optimization problems discussed in Section \ref{distances}. For that, we utilize numerical algebraic geometry algorithms implemented in the freeware packages \texttt{Bertini} and \texttt{HC.jl}. 

The pipeline for computing the closest architectural design for each case is executed in three phases, which are discussed in the subsequent subsections.

\subsection{Ab-initio phase} \label{step1}
{We pick a random set of complex base points $\mathbf{M}^{\mathbb{C}}_i$ with respect to the 
fixed frame $\frak{F}_{0}$  and complex platform points $\mathbf{m}_i^{\mathbb{C}}$ with respect to the moving frame  $\frak{F}$ (for $i=1,\ldots, 5$).} Let us denote the squared distance function $D_j^2$ by $L_j$ for $j=0,1,2,3a,4,5a,5b,6,7$. Using this notation, the closest architecture singularity of case $k$
discussed in the previous section corresponds to the minimization of the function $L_k$ with $k=0,1,2,3a,3b,4,5a,5b,6,7,8,9$. Due to the algebraic nature of these functions, their partial derivatives with respect to the 
mentioned unknowns (including the Lagrange multipliers), always result in square systems $P_k$ of polynomial equations. 
In all equations of $P_k$ the coefficients are scaled roughly to have the same magnitude between $0$ and $1$ (cf.\ \cite[Sec.\ 7.7]{b9}) to maintain computational accuracy and reduce numerical errors during tracking.

\begin{enumerate}
\item {\textbf{Higher dimensional components:}}
First, we check if the solution set of $P_k$  contains any higher dimensional components by invoking $\mathtt{Track Type 1}$ (see \cite[Sec 8.2]{b9}) in the \texttt{Bertini} input file.

For $k=0,1,2,3a,3b,4,5a,6, 7$, we conclude that there does not exist any higher dimensional components. For case 5b, there is a one-dimensional component, which does not fulfill the assumption of case {5b}, as $\mathbf{M}^{'}_{1}=\mathbf{M}^{'}_{2}$ holds true. 
For cases 8 and 9, we are not able to comment on the existence of higher dimensional components as the computations exceed our computational limits\footnote{\label{note1}
The computations were performed in parallel using a total of 64 threads using AMD Ryzen 7 2700X, 3.7 GHz processor with 63\,973\,484 kB RAM.}.

\item 
{\textbf{Isolated solutions:}} Now we proceed to compute generic finite solutions for $k=0,1,2,3a,4,5a,5b,6,7$. As for these cases the number of unknowns and the polynomial degree is low, one can use multi-homogeneous homotopy for computing generic finite solutions with \texttt{Bertini}. Lastly, in Table~\ref{data} we present the root count for these cases by eliminating the non-valid solutions; i.e.\ solutions already covered by  previous cases.  

Note that for all these cases we were also able to solve the system of equations 
by using Gr\"obner basis method in \texttt{Maple}. The obtained solutions match with the root count provided by \texttt{Bertini}. By performing this verification, we confirmed that all solutions were tracked.

For case 3b we employed a multi-homogeneous homotopy and achieved a ``perfect tracking'', i.e.\ no path failures or warnings from \texttt{Bertini}. The valid root count for this case is also given in  Table~\ref{data}.

{Therefore, we only remain with the discussion of the architecture singular cases 8 and 9, which is done in the following section.}
\end{enumerate}

\begin{table*}[]
\centering
\caption{Finite solution root count and path failures for the ab-initio phase for cases 8 and 9.}
\begin{tabular}{|l||llll||llll|}
\hline
 &
  \multicolumn{4}{c||}{Non-planar (Cartesian coordinates)} &
  \multicolumn{4}{c|}{Planar (Isotropic coordinates)} \\ \hline
 &
  \multicolumn{2}{c|}{case 8 (\texttt{HC.jl})} &
  \multicolumn{2}{c||}{ case 9 (\texttt{HC.jl})} &
  \multicolumn{2}{c|}{case 8 (\texttt{HC.jl}) } &
  \multicolumn{2}{c|}{case 9 (\texttt{HC.jl})} \\ \hline
Run &
  \multicolumn{1}{l|}{\begin{tabular}[c]{@{}l@{}}$\#$ finite sol-\\utions over $\mathbb{C}$\end{tabular}} &
  \multicolumn{1}{l|}{\begin{tabular}[c]{@{}l@{}}$\#$ path\\ failures\end{tabular}} &
  \multicolumn{1}{l|}{\begin{tabular}[c]{@{}l@{}} $\#$ finite sol-\\utions over $\mathbb{C}$\end{tabular}} &
  \begin{tabular}[c]{@{}l@{}}$\#$ path\\ failures\end{tabular} &
  \multicolumn{1}{l|}{\begin{tabular}[c]{@{}l@{}} $\#$ finite sol-\\utions over $\mathbb{C}$\end{tabular}} &
  \multicolumn{1}{l|}{\begin{tabular}[c]{@{}l@{}} $\#$ path \\ failures\end{tabular}} &
  \multicolumn{1}{l|}{\begin{tabular}[c]{@{}l@{}} $\#$ finite sol-\\utions  over $\mathbb{C}$ \end{tabular}} &
  \begin{tabular}[c]{@{}l@{}}$\#$ path \\ failures\end{tabular} \\ \hline \hline
1 &
  \multicolumn{1}{l|}{380} &
  \multicolumn{1}{l|}{4 768} &
  \multicolumn{1}{l|}{3 062} &
  267 446 &
  \multicolumn{1}{l|}{356} &
  \multicolumn{1}{l|}{1 950} &
  \multicolumn{1}{l|}{2 691} &
  64 333 \\ \hline
2 &
  \multicolumn{1}{l|}{380} &
  \multicolumn{1}{l|}{5 020} &
  \multicolumn{1}{l|}{3 068} &
  300 446 &
  \multicolumn{1}{l|}{355} &
  \multicolumn{1}{l|}{1 463} &
  \multicolumn{1}{l|}{2 690} &
  61 863 \\ \hline
3 &
  \multicolumn{1}{l|}{380} &
  \multicolumn{1}{l|}{5 131} &
  \multicolumn{1}{l|}{3 053} &
  309 585 &
  \multicolumn{1}{l|}{356} &
  \multicolumn{1}{l|}{1 544} &
  \multicolumn{1}{l|}{2 690} &
  61 863 \\ \hline
4 &
  \multicolumn{1}{l|}{379} &
  \multicolumn{1}{l|}{5 506} &
  \multicolumn{1}{l|}{3 064} &
  288 559 &
  \multicolumn{1}{l|}{354} &
  \multicolumn{1}{l|}{1 433} &
  \multicolumn{1}{l|}{2 684} &
  61 863 \\ \hline
5 &
  \multicolumn{1}{l|}{379} &
  \multicolumn{1}{l|}{7 725} &
  \multicolumn{1}{l|}{3 062} &
  293 622  &
  \multicolumn{1}{l|}{355} &
  \multicolumn{1}{l|}{1 433} &
  \multicolumn{1}{l|}{2 691} &
  65 567 \\ \hline
\end{tabular}
\label{finaltable}
\end{table*}

\subsubsection{Discussion on cases 8 and 9}
We discuss these cases for designs with non-planar as well as planar base (cf.\ Remark \ref{rmk:planar}).

It can be observed from Table~\ref{data} that for both cases the degree of $L_k$ and their partial derivatives for the listed number of unknowns result in large polynomial systems for non-planar as well as planar designs. To efficiently track all the solutions, including the singular solutions, using a multi-homogeneous homotopy method, it is essential to identify the optimal grouping that minimizes the number of paths to be tracked, with respect to the total number of unknowns listed in Table~\ref{data}. However, as mentioned in (cf. \cite[pg. 72]{b9}), it is evident that finding the best group is NP-hard~\cite{21}, making the search for it as well as the use of multi-homogeneous homotopy computationally not feasible for these cases (taking our limited computational resources{\footref{note1} into account)}.

On the other hand, there is a possibility of using the regeneration algorithm \cite{b20}, which is less computationally demanding than multi-homogeneous homotopy, but it has the drawback that singular solutions are not covered (cf.\ \cite[pg. 89]{b9}).
One possible way to overcome this limitation and find both singular and non-singular solutions for cases 8 and 9 is to employ polyhedral homotopy, which is implemented in \texttt{HC.jl}\footnote{Polyhedral homotopy is not available in \texttt{Bertini 1.6v}.} \cite{b11}.

\paragraph{Designs with non-planar base:}
{
The BKK bound, which gives the number of paths to be tracked, for case 8  is 318\,312  and for case 9 it is  6\,203\,620, respectively. 
} 
{For cases 8 and 9, we encountered path failures (without considering paths that go to infinity) during tracking with \texttt{HC.jl}. The number of path failures for 5 runs and the generic finite solution root count 
is given in Table~\ref{finaltable}. By considering the run that resulted in the maximum number of solutions, we present in Table \ref{data} the number of valid solutions by eliminating the those already covered by previous cases.}

\paragraph{Design with planar base:}
{
The BKK bound for cases 8 and 9 is 107\,312, and 1,827\,040, respectively. 
But we can further reduce these bounds by using isotropic coordinates as proposed by Wampler \cite[Sec.\ 4.1]{b10}.  The mapping between Cartesian coordinates $(x,y)$ and isotopic coordinates $(p,\bar{p})$ is an invertible linear map: }
\begin{equation}\label{eq:cart_iso}
    (p, \bar{p})=(x+yi, x-yi), \quad (x,y)=\left(\frac{p+\bar{p}}{2}, \frac{p-\bar{p}}{2i}\right).
\end{equation}
{Hence, anything formulated in one set of coordinates can easily be rewritten in the other set by applying the substitutions of Eq.\ (\ref{eq:cart_iso}). For example, the distance function of cases 8 and 9 expressed in isotropic coordinates of the base anchor points reads as follows:}
\begin{equation}
    \Tilde{D}={\frac{1}{10}\sum_{i=1}^{5}\left[{(p_{i}-p'_{i})(\bar{p}_{i}- \bar{p}'_{i})}+(r_{i}-r'_{i})^2\right]}.
    \label{wampler}
\end{equation}
Resorting to this formulation, the BKK bound for case 8 and case 9 drops to 35\,976, and 598\,812, respectively.
As the usage of isotropic coordinates has significantly reduced these bounds, we utilize this approach along with polyhedral homotopy implemented in \texttt{HC.jl}.
 The number of path failures for 5 runs is summarized in Table ~\ref{finaltable}. Moreover, the valid generic finite solution root count is given in  Table~\ref{data}.

\begin{rmk}\label{rmk:proof}
Please note that path failures in cases 8 and 9 may occur when the starting system contains more zeros than the target system. In such instances, overhead paths are likely to fail. However, it is impossible to determine definitively whether the start system contains more zeros than the target system in \texttt{HC.jl} ~\cite{b19}. 
This can be supported by considering case 3b as an example. We consider a design with a planar base using Cartesian coordinates to track generic finite solutions with multi-homogeneous homotopy implemented in \texttt{Bertini} and polyhedral homotopy implemented in \texttt{HC.jl}. Table~\ref{pathfailures}  of Appendix~\ref{datacoordinates} provides a summary of path failures and root counts for $5$ runs. The \texttt{End game options} and \texttt{Tracker options} used for the \texttt{HC.jl} runs 
are given in Tables \ref{settings1} and \ref{settings2} of Appendix~\ref{datacoordinates}. 
The configuration settings used for the \texttt{Bertini} runs are SECURITYLEVEL:1, TRACKTolBEFOREEG:1e-8 and
TRACKTolDURINGEG:1e-8.
\hfill $\diamond$
\end{rmk}

\subsection{User-defined homotopy phase}\label{user} Based on the user input, which are the real base points $\mathbf {M}_{i}$ and the corresponding real platform points $\mathbf {m}_{i}$, 
we establish a linear homotopy path between $\mathbf{M}^{\mathbb{C}}_i$ and $\mathbf {M}_{i}$ as well as  $\mathbf{m}^{\mathbb{C}}_i$ and $\mathbf {m}_{i}$  
by:
\begin{equation}\label{var1}
{
\mathbf{U}_{i}} ={h}\mathbf{M}^{\mathbb{C}}_i +(1-{h})\mathbf{M}_{\ell_i}, \quad
\mathbf{u}_{i} ={h}\mathbf{m}^{\mathbb{C}}_i +(1-{h})\mathbf{m}_{\ell_i},
\end{equation}
for $i=1,\ldots, 5$ and track the finite solutions of the {\it ab-initio phase}, while the {homotopy} parameter $h$ reduces from $1$ to $0$.  
To avoid infinitely long paths we constructed user-defined homotopies on product spaces \cite[Sec.\ 7.5.2]{b9}. Note that in Eq.\ (\ref{var1}) the indices $\ell_{1},\ldots ,\ell_{5}$  are pairwise distinct with $\ell_{1},\ldots ,\ell_{5}\in\left\{1,\ldots,5\right\}$ 
which allows in general $5!=120$ combinatorial cases, but we do not have to consider all of them as multiple combinations end up in the same 
function $L_k$. The essential number of combinatorial cases, for which the user-defined homotopy has to be repeated in each architecture singular case, is also given in Table \ref{data}. 

In more detail, we replace $\mathbf{M}^{\mathbb{C}}_i$ and $\mathbf{m}^{\mathbb{C}}_i$ with $\mathbf{U}_{i}$ and $\mathbf{u}_{i}$ from Eq.\ (\ref{var1}) in the distance functions of the respective design cases outlined in Section~\ref{distances}. Again, their partial derivatives to the mentioned unknowns give rise to the user-defined homotopy system of polynomial equations, involving a homotopy parameter $h$. This system is then handed over to \texttt{Bertini} to trace the finite isolated solutions of the ab-initio phase given in Table~\ref{data}. If
there are path failures during this tracking, one has to tighten the tracking tolerances and recompute for the failed paths.
By varying the indices of base and platform coordinates we obtain the combinations listed in Table~\ref{data}. 

\subsection{Post-processing phase} \label{post}
{
In this phase, the obtained real solutions are used as input for \texttt{Maple}. For the corresponding design combination, the distance function is evaluated to identify the global minimizer.}

Finally, it should be noted that the generic finite solutions of the {\it ab-initio phase} need only to be computed once for each of the different optimization problems. All the generic finite solutions are computed in floating arithmetic up to $16$ digits and they are  provided as
supplementary material~\cite{b18} as well as the 
corresponding complex coordinates $\mathbf{M}_i^{\mathbb{C}}$ and  $\mathbf{m}_i^{\mathbb{C}}$. 
Using this data one only has to run Sections \ref{user} and \ref{post} for the computation of the architecture singularity distance.

\section{Results and discussion}\label{results}
In the following subsections, we present two numerical examples; the first studied design has a planar base (cf.\ Section \ref{ex:planar}) and the second a non-planar one  (cf.\ Section \ref{ex:nonplanar}). We employ the computational procedure outlined in Section~\ref{sec:procedure1} and conduct user-defined homotopies from Section \ref{user}, along with post-processing as described in Section~\ref{post}, for all possible design combinations.

\subsection{Example with planar base}\label{ex:planar}
    
{
{We use the following numerical example provided in  \cite[Sec.\ 4.1]{b3} as input for the design of our linear pentapod with planar base:}
$\mathbf{M}_{1}=(0,2,0)_{\frak{F}_{0}}$, $\mathbf{M}_{2}=(\frac{-3}{2},\frac{9}{4},0)_{\frak{F}_{0}}$, 
$\mathbf{M}_{3}=(-3, 1, 0)_{\frak{F}_{0}}$, $\mathbf{M}_{4}=(-1,0,0)_{\frak{F}_{0}}$, 
$\mathbf{m}_{1}=(0,0,0)_{\frak{F}}$,  $\mathbf{m}_{2}=(1,0,0)_{\frak{F}}$, $\mathbf{m}_{3}=(2, 0, 0)_{\frak{F}}$, $\mathbf{m}_{4}=(3,0,0)_{\frak{F}}$, 
$\mathbf{m}_{5}=(4,0,0)_{\frak{F}}$.
}

\begin{figure*}[]
\centering
\begin{subfigure}[b]{0.23\textwidth}
    \centering
    \begin{overpic}[width=\linewidth]{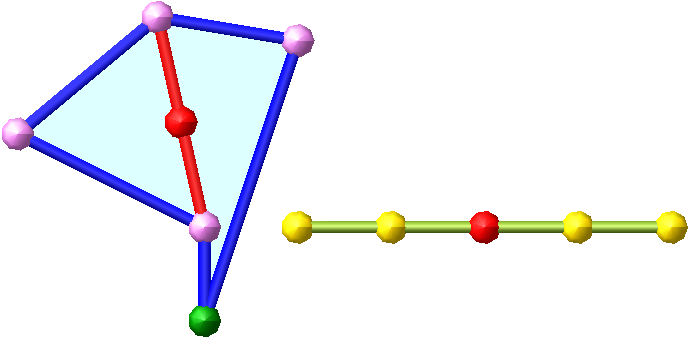}
        \begin{small}
      \put(17,11.5){$\mathbf{M}_{4}$}
      \put(7,28){$\mathbf{M'}_{2,4}$}
      \put(15,50){$\mathbf{M}_{2}$}
       \put(60,22){$\mathbf{m}'_{2}=\mathbf{m}'_{4}$}
        \put(55,8){$\mathbf{m}_{2}$}
         \put(80,8){$\mathbf{m}_{4}$}
         \put(68,8){$\mathbf{m}_{3}$}
        \end{small}
    \end{overpic}
    \caption{Case 0}
\end{subfigure}\quad\quad\quad\quad
\begin{subfigure}[b]{0.23\textwidth}
    \centering
    \begin{overpic}[width=\linewidth]{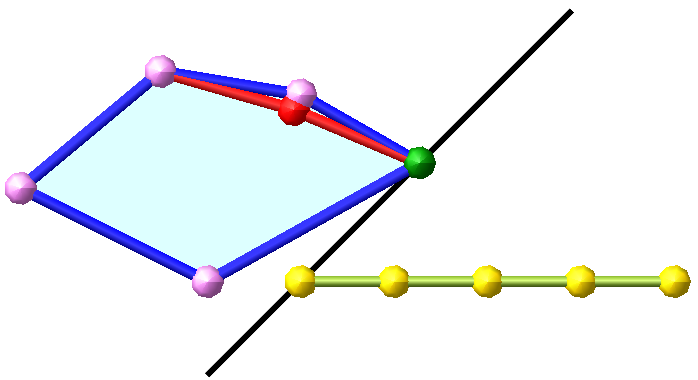}
        \begin{small}
        \put(63,28){$\mathbf{M}_{5}$}
       \put(20,50){$\mathbf{M}_{2}$}
        \put(40,45){$\mathbf{M}_{1}$}
         \put(30,30){$\mathbf{M}'_{1,2,5}$}
        \end{small}
    \end{overpic}
    \caption{Case 1}
\end{subfigure}\quad\quad\quad\quad
\begin{subfigure}[b]{0.23\textwidth}
    \centering
    \begin{overpic}[width=\linewidth]{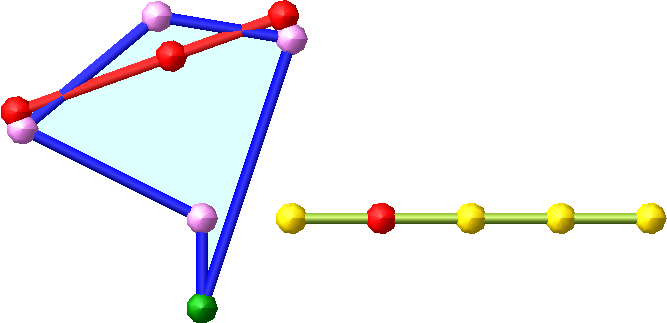}
        \begin{small}
        \put(20,50){$\mathbf{M}_{2}$}
        \put(46,38){$\mathbf{M}_{1}$}
        \put(42,50){$\mathbf{M}'_{1}$}
        \put(20,30){$\mathbf{M}'_{2}$}
         \put(0,20){$\mathbf{M}_{3}$}
         \put(-2,38){$\mathbf{M}'_{3}$}
          \put(50,22){$\mathbf{m}'_{1,2,3}$}
           \put(40,8){$\mathbf{m}_{1}$}
            \put(55,8){$\mathbf{m}_{2}$}
             \put(68,8){$\mathbf{m}_{3}$}
        \end{small}
    \end{overpic}
    \caption{Case 2}
\end{subfigure}

\vspace{1em}  

\begin{subfigure}[b]{0.23\textwidth}
    \centering
    \begin{overpic}[width=\linewidth]{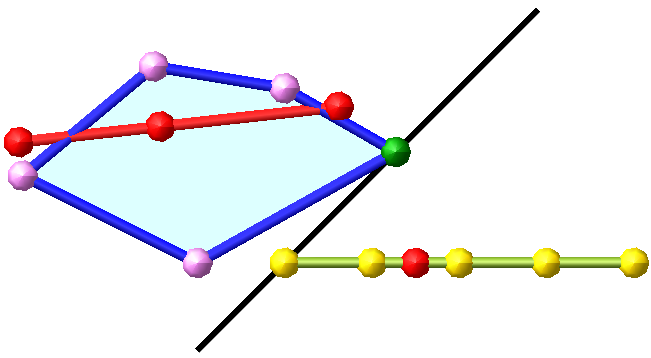}
        \begin{small}
     \put(52,43){$\mathbf{M}'_{1,5}$}
      \put(23,26){$\mathbf{M}'_{2}$}
      \put(-3,38){$\mathbf{M}'_{3}$}
      \put(55,20){$\mathbf{m}'_{2,3}$}
       \put(63,28){$\mathbf{M}_{5}$}
        \put(40,46){$\mathbf{M}_{1}$}
        \put(0,20){$\mathbf{M}_{3}$}
        \put(20,50){$\mathbf{M}_{2}$}
        \end{small}
    \end{overpic}
   
    \caption{Case 3a}
\end{subfigure}\quad\quad\quad\quad
\begin{subfigure}[b]{0.23\textwidth}
    \centering
    \begin{overpic}[width=\linewidth]{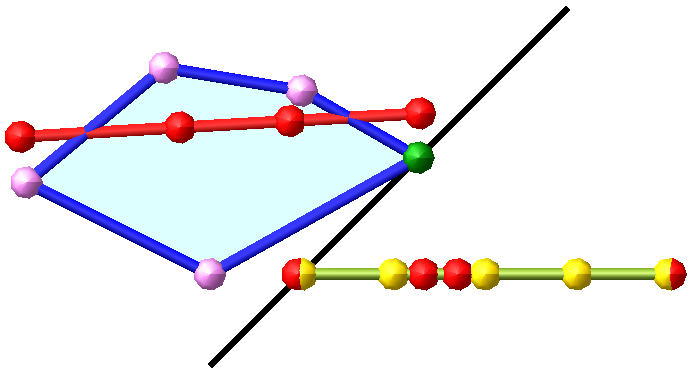}
        \begin{small}
             \put(42,45){$\mathbf{M}_{1}$}
    \put(39,28){$\mathbf{M}'_{1}$} 
    \put(-3,38){$\mathbf{M}'_{3}$}
     \put(-3,20){$\mathbf{M}_{3}$}    \put(63,26){$\mathbf{M}_{5}$}
     \put(55,42){$\mathbf{M}'_{5}$}
     \put(20,50){$\mathbf{M}_{2}$}
     \put(20,27){$\mathbf{M}'_{2}$}
      \put(40,6){$\mathbf{m}'_{1}$}
      \put(57,6){$\mathbf{m}'_{2}$}
      \put(61,19){$\mathbf{m}'_{3}$}
       \put(93,6){$\mathbf{m}'_{5}$}
        \end{small}
    \end{overpic}
 
    \caption{Case 3b}
\end{subfigure}\quad\quad\quad\quad
\begin{subfigure}[b]{0.23\textwidth}
    \centering
    \begin{overpic}[width=\linewidth]{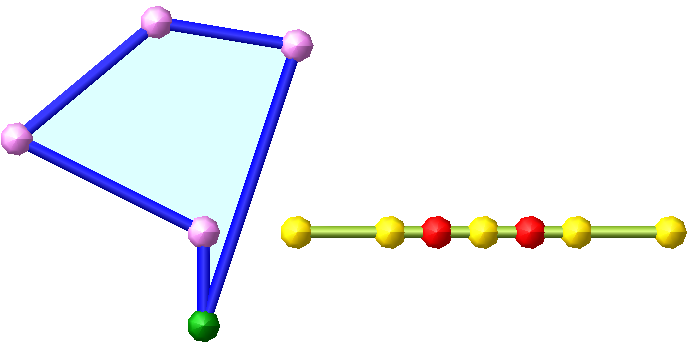}
        \begin{small}
        \put(53,22){$\mathbf{m}'_{1,2,3,4}$}
        \put(73,8){$\mathbf{m}'_{2,3,4,5}$}
        \end{small}
    \end{overpic}
  
    \caption{Case 4}
\end{subfigure}

\vspace{1em}  

\begin{subfigure}[b]{0.23\textwidth}
    \centering
    \begin{overpic}[width=\linewidth]{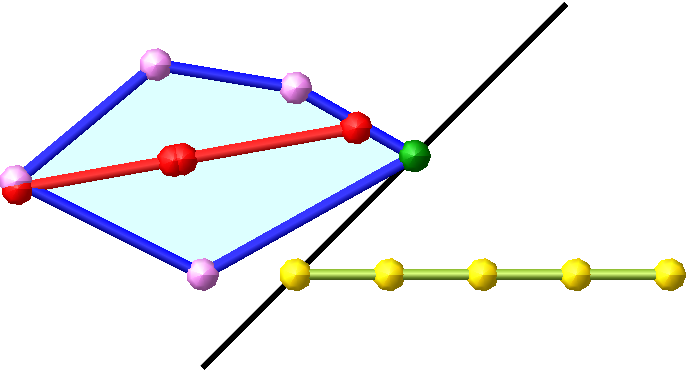}
        \begin{small}
    \put(63,28){$\mathbf{M}_{5}$}
    \put(53,40){$\mathbf{M}'_{1}=\mathbf{M}'_{5}$}
    \put(43,45){$\mathbf{M}_{1}$}
    \put(20,48){$\mathbf{M}_{2}$}
    \put(0,18){$\mathbf{M}'_{3}$}
    \put(-3,36){$\mathbf{M}_{3}$}
    \put(25,36){$\mathbf{M}'_{2}$}
    \put(22,8){$\mathbf{M}_{4}$}
    \put(21,23){$\mathbf{M}'_{4}$}
        \end{small}
    \end{overpic}
   \caption{Case 5a}
\end{subfigure}\quad\quad\quad\quad
\begin{subfigure}[b]{0.23\textwidth}
    \centering
    \begin{overpic}[width=\linewidth]{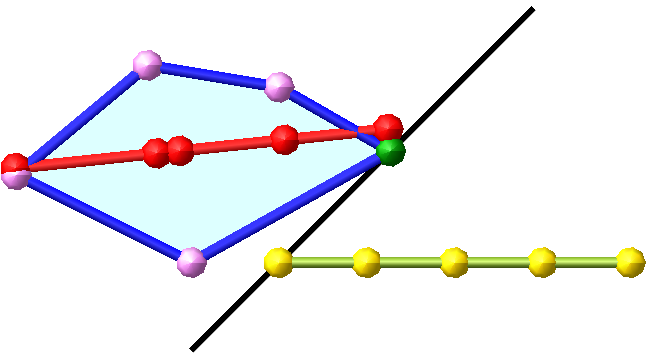}
        \begin{small}
            \put(52,40){$\mathbf{M}'_{5}$}
    \put(0,18){$\mathbf{M}_{3}$}
     \put(-2,35){$\mathbf{M}'_{3}$}
        \put(25,36){$\mathbf{M}'_{2}$}
        \put(20,50){$\mathbf{M}_{2}$}
        \put(21,23){$\mathbf{M}'_{4}$}
        \put(63,28){$\mathbf{M}_{5}$}
        \end{small}
    \end{overpic}
  
    \caption{Case 5b}
\end{subfigure}\quad\quad\quad\quad
\begin{subfigure}[b]{0.23\textwidth}
    \centering
    \begin{overpic}[width=\linewidth]{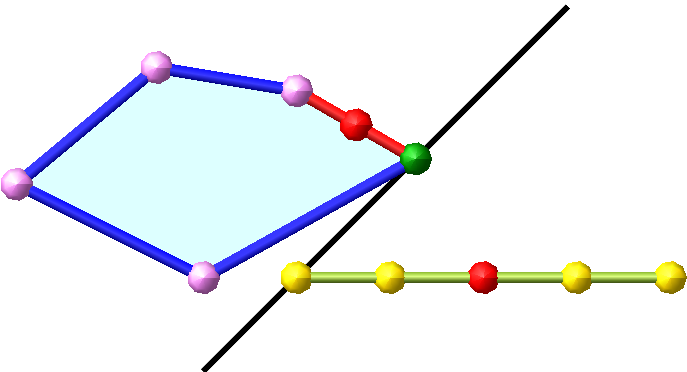}
        \begin{small}
           \put(63,28){$\mathbf{M}_{5}$}
       \put(40,45){$\mathbf{M}_{1}$}
        \put(50,39.5){$\mathbf{M'}_{1,5}$}
         \put(68,6){$\mathbf{m}'_{1,2,3}$}
        \end{small}
    \end{overpic}
   \caption{Case 6}
\end{subfigure}

\vspace{1em}  

\begin{subfigure}[b]{0.23\textwidth}
    \centering
    \begin{overpic}[width=\linewidth]{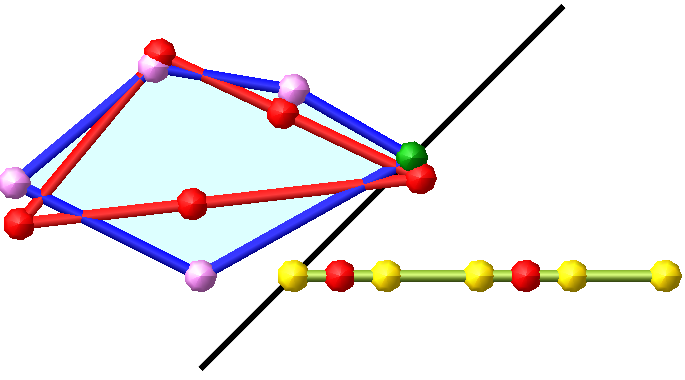}
        \begin{small}
            \put(54,8)

              \put(61,22){$\mathbf{M'}_{5}$}
            \put(63,31){$\mathbf{M}_{5}$}
            \put(20,50){$\mathbf{M}'_{2}$}
        \put(40,45){$\mathbf{M}_{1}$}
        \put(38,30){$\mathbf{M}'_{1}$}
      \put(17,11.5){$\mathbf{M}_{4}$}
     \put(-4,34) {$\mathbf{M}_{3}$}
    \put(0,13) {$\mathbf{M}'_{3}$}
   \put(20,32){$\mathbf{M}'_{4}$}
   \put(73,6){$\mathbf{m}'_{3,4}$}
   \put(43,6){$\mathbf{m}'_{1,2}$}
        \end{small}
    \end{overpic}
    \caption{Case 7}
\end{subfigure}\quad\quad\quad\quad
\begin{subfigure}[b]{0.23\textwidth}
    \centering
    \begin{overpic}[width=\linewidth]{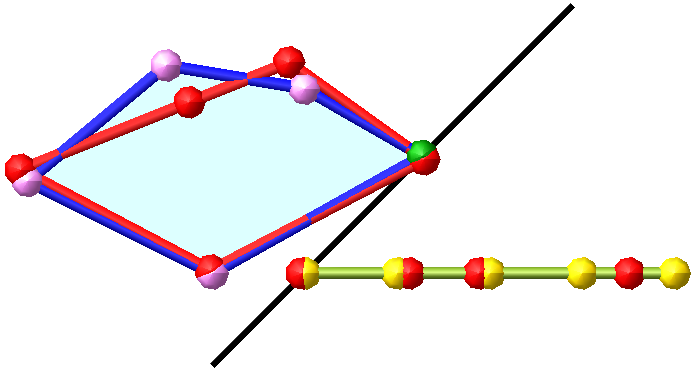}
        \begin{small}
            \put(40,6){$\mathbf{m}'_{1}$}
            \put(54,6){$\mathbf{m}'_{2}$}
             \put(64,6){$\mathbf{m}'_{3}$}
    \put(82,20){$\mathbf{m}'_{4,5}$}
    \put(61,22){$\mathbf{M'}_{5}$}
      \put(63,31){$\mathbf{M}_{5}$}
       \put(46,42){$\mathbf{M}_{1}$}
        \put(42,50){$\mathbf{M}'_{1}$}
      \put(20,48){$\mathbf{M}_{2}$}
       \put(20,30){$\mathbf{M}'_{2}$}
        \put(0,18){$\mathbf{M}_{3}$}
     \put(-2,35){$\mathbf{M}'_{3}$}
      \put(22,6){$\mathbf{M}_{4}$}
      \put(22,21){$\mathbf{M}'_{4}$}
        \end{small}
    \end{overpic}
   \caption{Case 8}
\end{subfigure}\quad\quad\quad\quad
\begin{subfigure}[b]{0.23\textwidth}
    \centering
    \begin{overpic}[width=\linewidth]{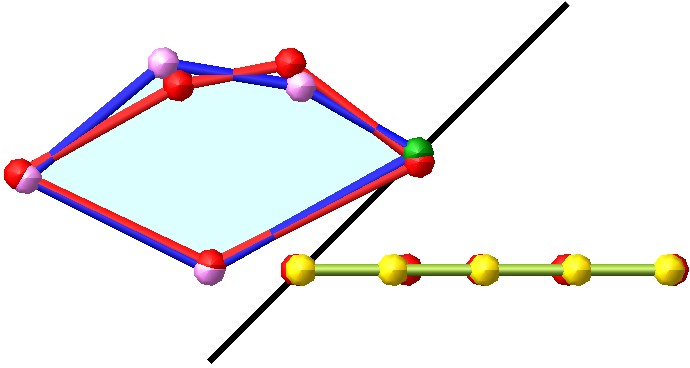}
        \begin{small}
          \put(40,6){$\mathbf{m}'_{1}$}
            \put(54,6){$\mathbf{m}'_{2}$}
             \put(65,6){$\mathbf{m}'_{3}$}
    \put(78,20){$\mathbf{m}'_{4}$}
    \put(92,20){$\mathbf{m}'_{5}$}
    \put(61,22){$\mathbf{M'}_{5}$}
      \put(64,30){$\mathbf{M}_{5}$}
       \put(46,42){$\mathbf{M}_{1}$}
        \put(42,50){$\mathbf{M}'_{1}$}
      \put(22,48){$\mathbf{M}_{2}$}
       \put(23,31){$\mathbf{M}'_{2}$}
        \put(0,18){$\mathbf{M}_{3}$}
     \put(-2,35){$\mathbf{M}'_{3}$}
      \put(22,6){$\mathbf{M}_{4}$}
      \put(22,21){$\mathbf{M}'_{4}$}
        \end{small}
    \end{overpic}
  \caption{Case 9}
\end{subfigure}
\caption{Illustration of architectural singular designs for $t=0.383206$. For case 2, the equation of the line of regression equals $-0.33882567x+0.94084916y-2.15472456=0$. For case 5b, the line of  regression is given by  $-0.104481452x+0.99452683y-1.3854265686=0$. For case 4, it should be noted that the design combination $\mathbf{m}_{1,2,3,4}$ and $\mathbf{m}_{2,3,4,5}$ result in the same minimum distance. For the coordinates of the  base and platform points of the closest architectural singular designs we refer to Tables \ref{basecoordinates} and \ref{platformcoordinates} of Appendix~\ref{datacoordinates}.}
\label{geometricdesigns}
\end{figure*}

\noindent {For our study, we vary the base point $\mathbf{M}_{5}$ (green point in Fig.\ \ref{geometricdesigns}) along a line (black line in Fig.\ \ref{geometricdesigns}) in dependence of the design parameter $t$; i.e.\ }
\begin{equation}
 \mathbf{M}_{5}=\mathbf{B}+t\frac{\mathbf{B}-\mathbf{M}_{5}^*}{||\mathbf{B}-\mathbf{M}_{5}^*||},  
\end{equation}
where $\mathbf{M}_{5}^*=(-1, -1, 0)_{\frak{F}{0}}$ is the initial location of the $5^{th}$ base point according to \cite[Sec.\ 4.1]{b3} and $\mathbf{B}=(1,1,0)_{\frak{F}{0}}$.
We restrict $t$ to the interval $[2\sqrt{2}, -2\sqrt{2}]$ and discretize it into 45 equidistant points for the evaluation of the best design within this one-dimensional set of pentapods.

\begin{figure*}[]
\begin{multicols}{4}
    \begin{overpic}
    [width=\linewidth]{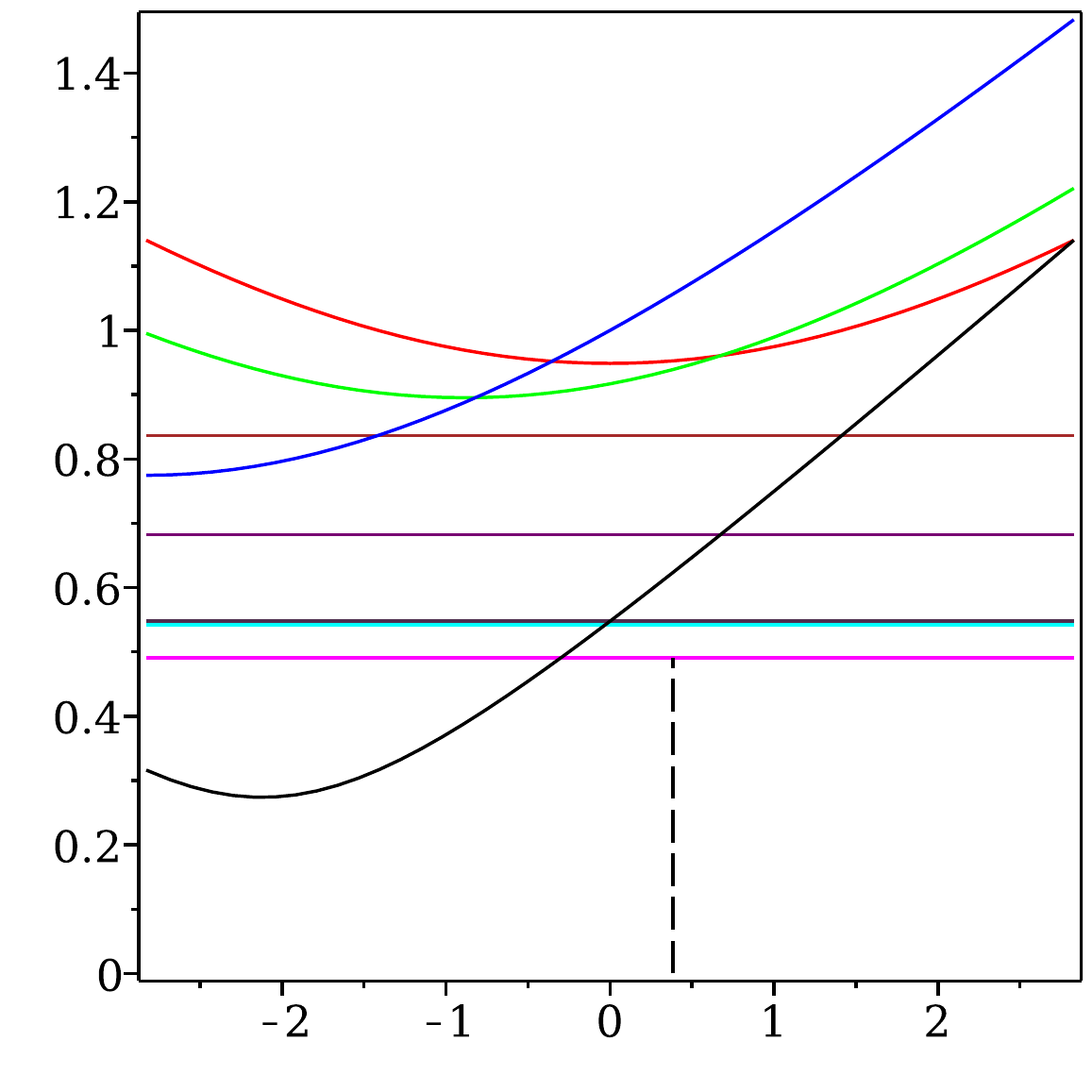}
    \put(45,-6){(a) Case 0}
     \put(0,60){\makebox(0,0){\rotatebox{90}{Distance}}}
     \put(92,3){$t$}
  \end{overpic}

\begin{overpic}
     [width=\linewidth]{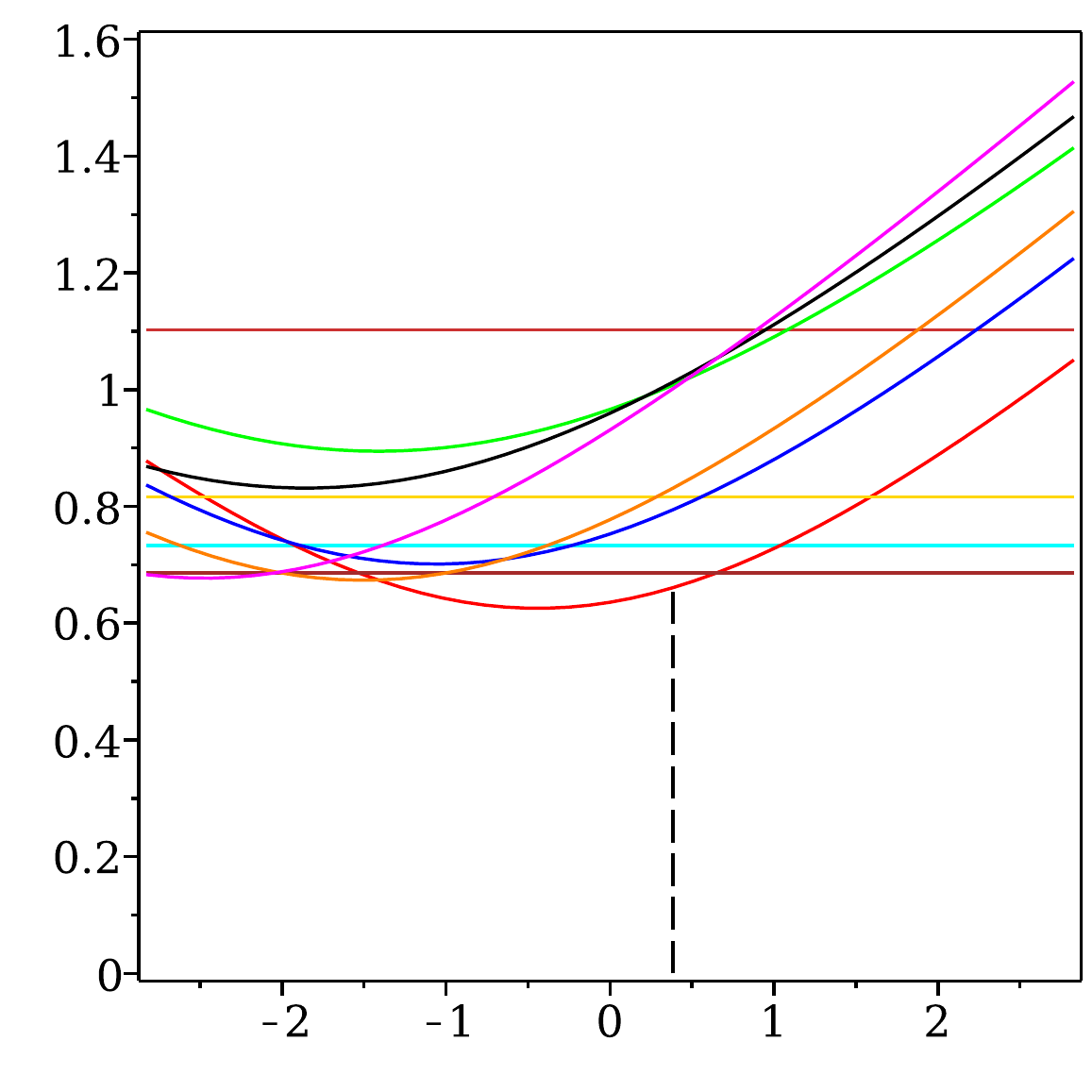}
      \put(45,-6){(b) Case 1}
      \put(0,60){\makebox(0,0){\rotatebox{90}{Distance}}}
       \put(92,3){$t$}
\end{overpic}

\begin{overpic}
     [width=\linewidth]{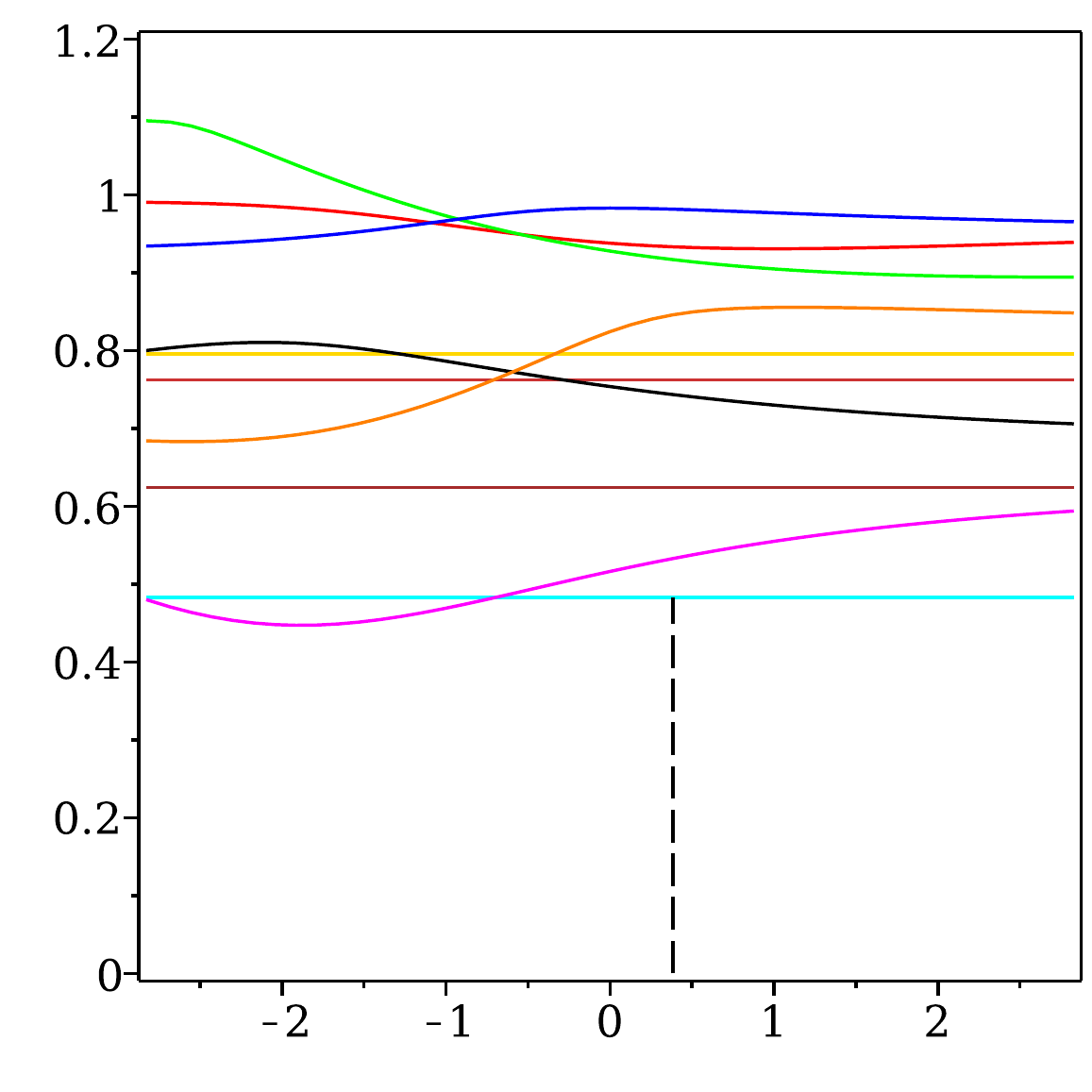}
      \put(45,-6){(c) Case 2}
       \put(0,60){\makebox(0,0){\rotatebox{90}{Distance}}}
        \put(92,3){$t$}
     \end{overpic}
     
\begin{overpic}
    [width=\linewidth]{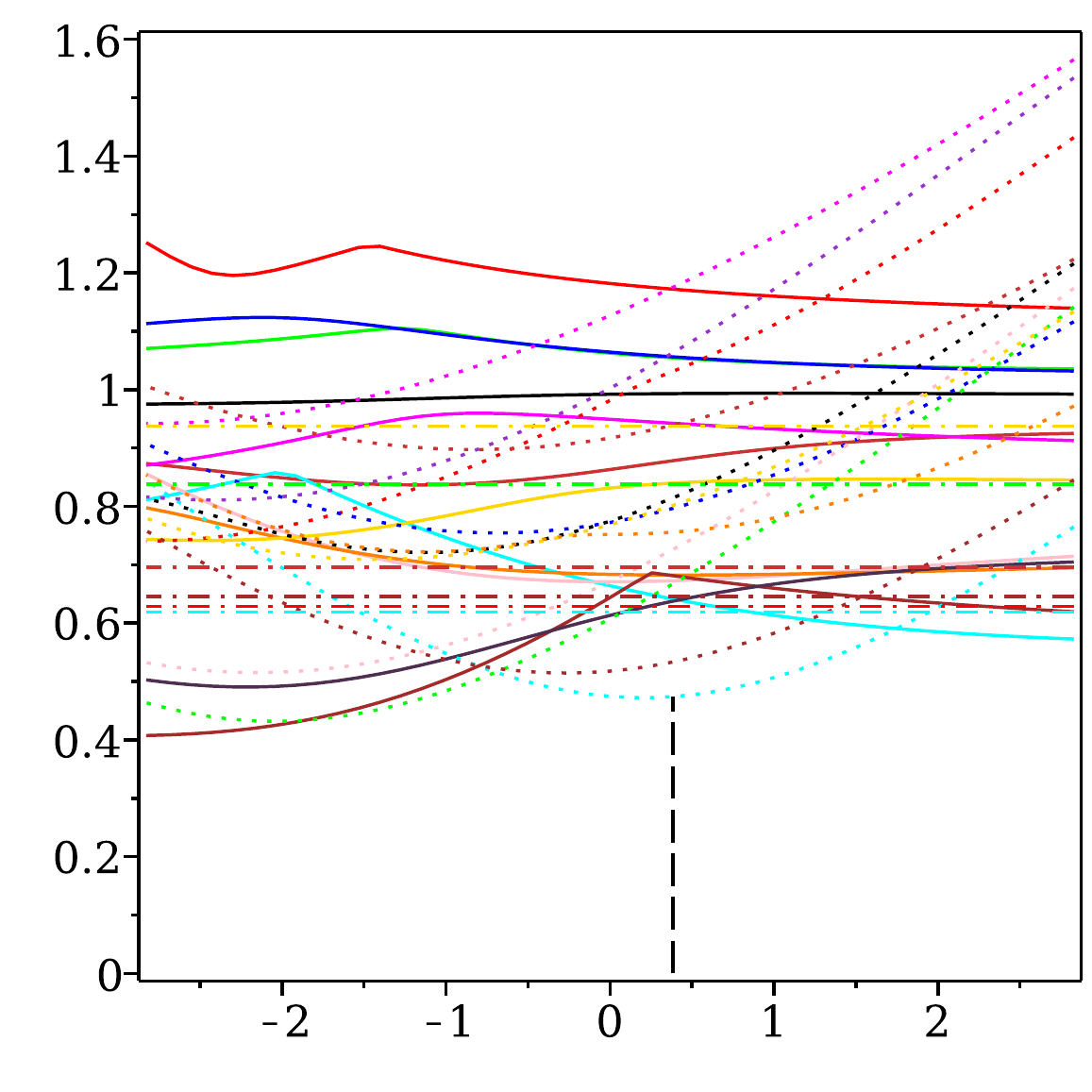}
     \put(45,-6){(d) Case 3a}
     \put(0,60){\makebox(0,0){\rotatebox{90}{Distance}}}
        \put(92,3){$t$}
     \end{overpic}
\end{multicols}

\begin{multicols}{4}
    \begin{overpic}
    [width=\linewidth]{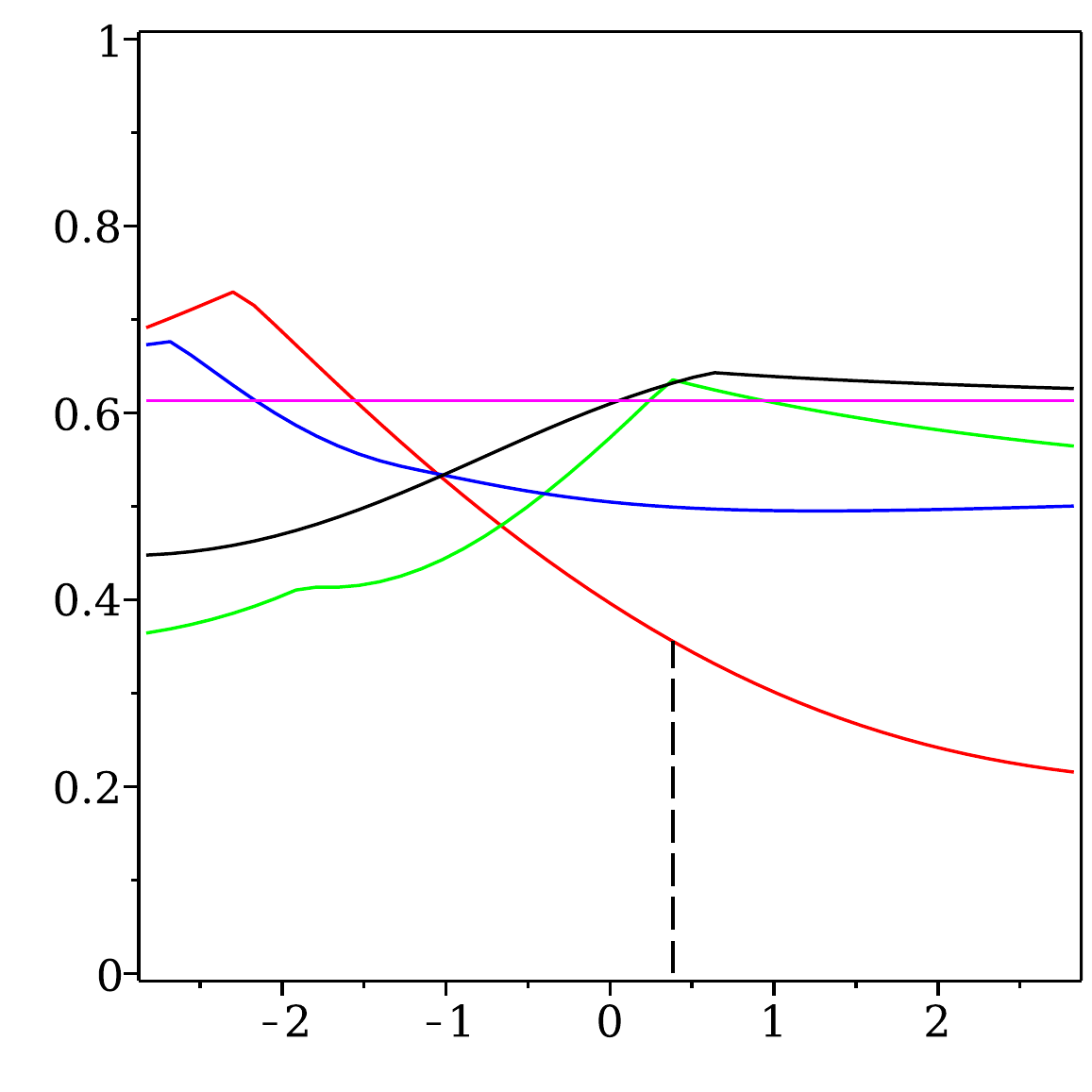} 
     \put(45,-6){(e) Case 3b}
     \put(0,60){\makebox(0,0){\rotatebox{90}{Distance}}}
        \put(92,3){$t$}
\end{overpic}

\begin{overpic}
    [width=\linewidth]{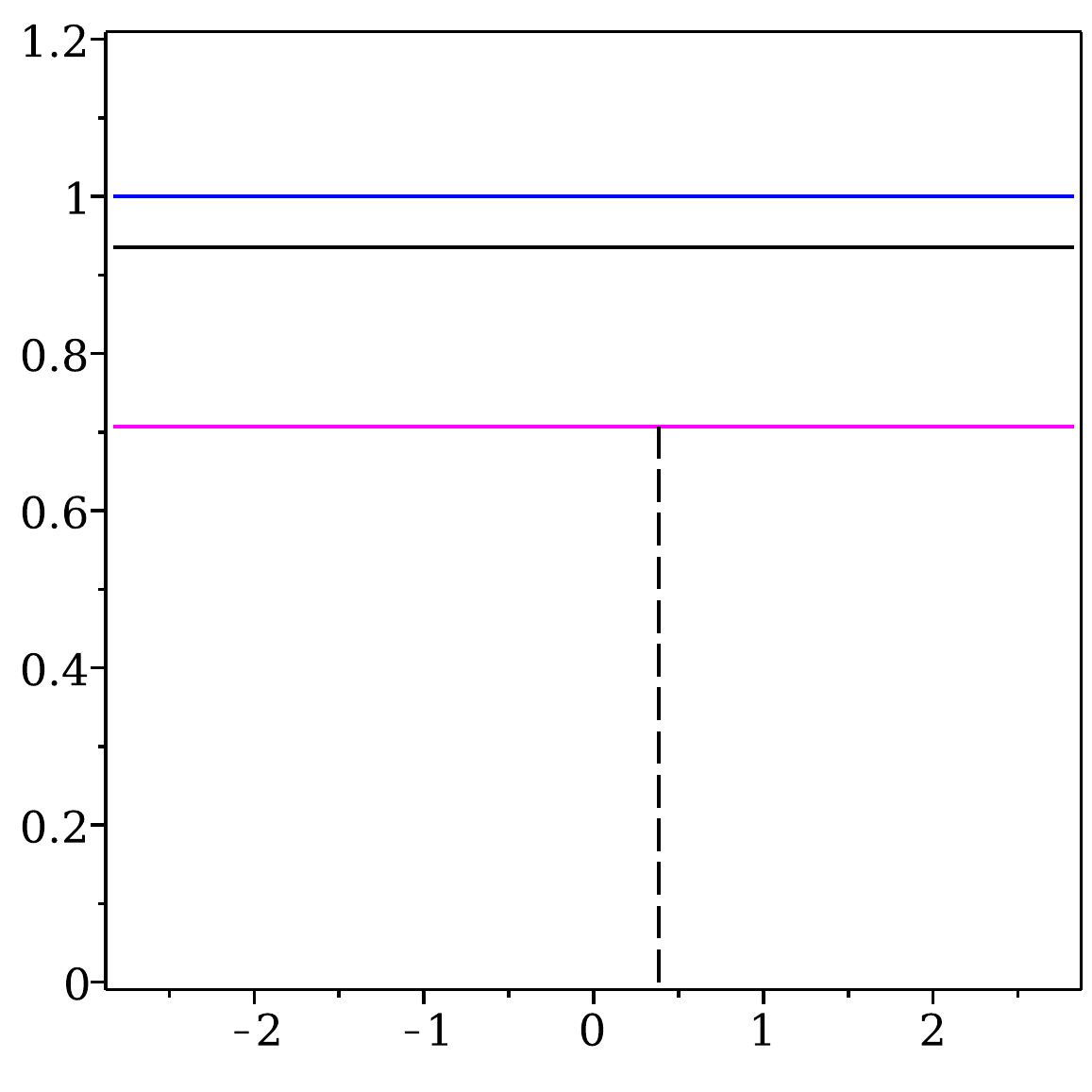}
    \put(45,-6){(f) Case 4}
    \put(-2,60){\makebox(0,0){\rotatebox{90}{Distance}}}
        \put(92,3){$t$}
\end{overpic}

\begin{overpic}
    [width=\linewidth]{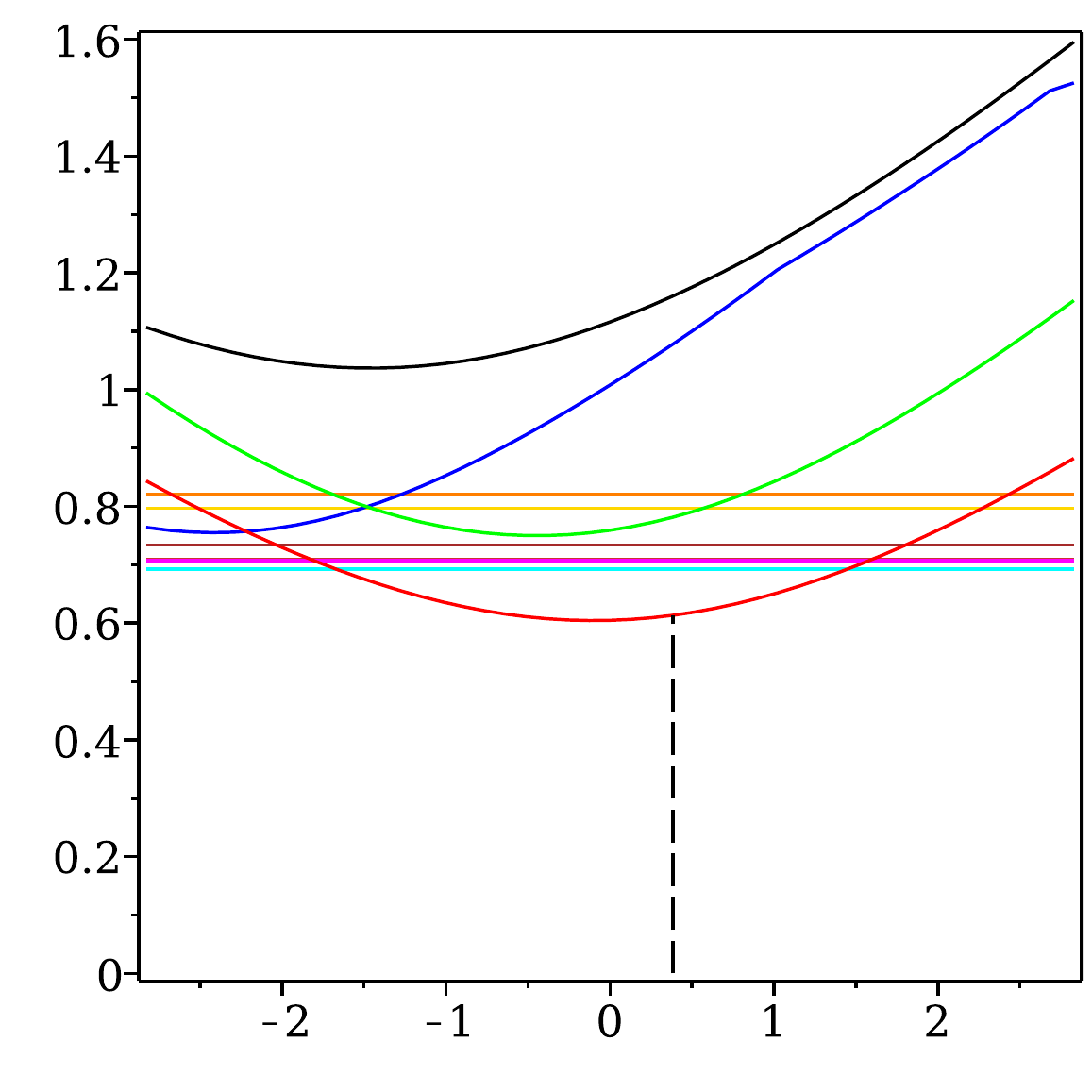}
    \put(40,-6){(g) Case 5a}
    \put(0,60){\makebox(0,0){\rotatebox{90}{Distance}}}
        \put(92,3){$t$}
  \end{overpic}\par
  
    \begin{overpic}
    [width=\linewidth]{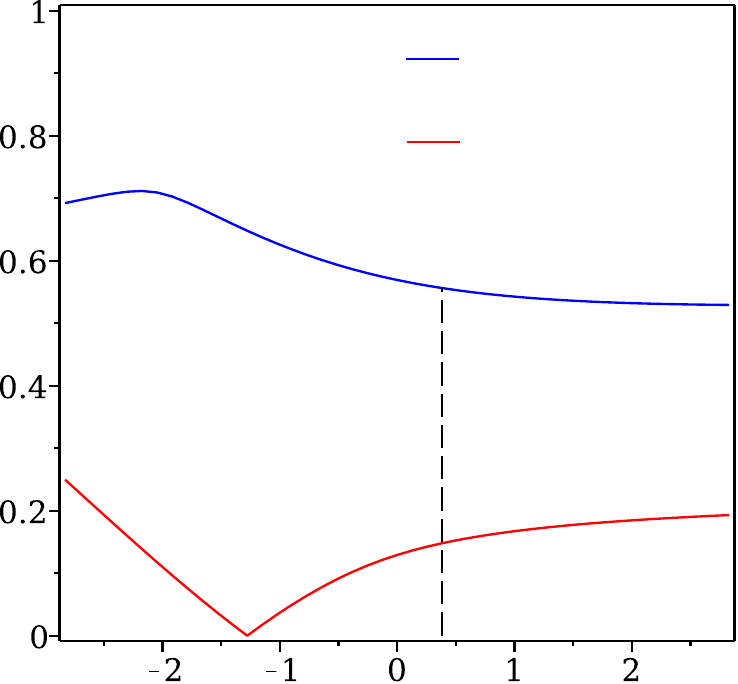}
    \put(33,-12){(h) Cases 5b and 9}
    \put(-5,60){\makebox(0,0){\rotatebox{90}{Distance}}}
        \put(92,-1){$t$}
         \put(65,70){$D_{9}$}
    \put(65,82) {$D_{5}$}
\end{overpic}\par
\end{multicols}

\begin{multicols}{4}
    \begin{overpic}
    [width=\linewidth]{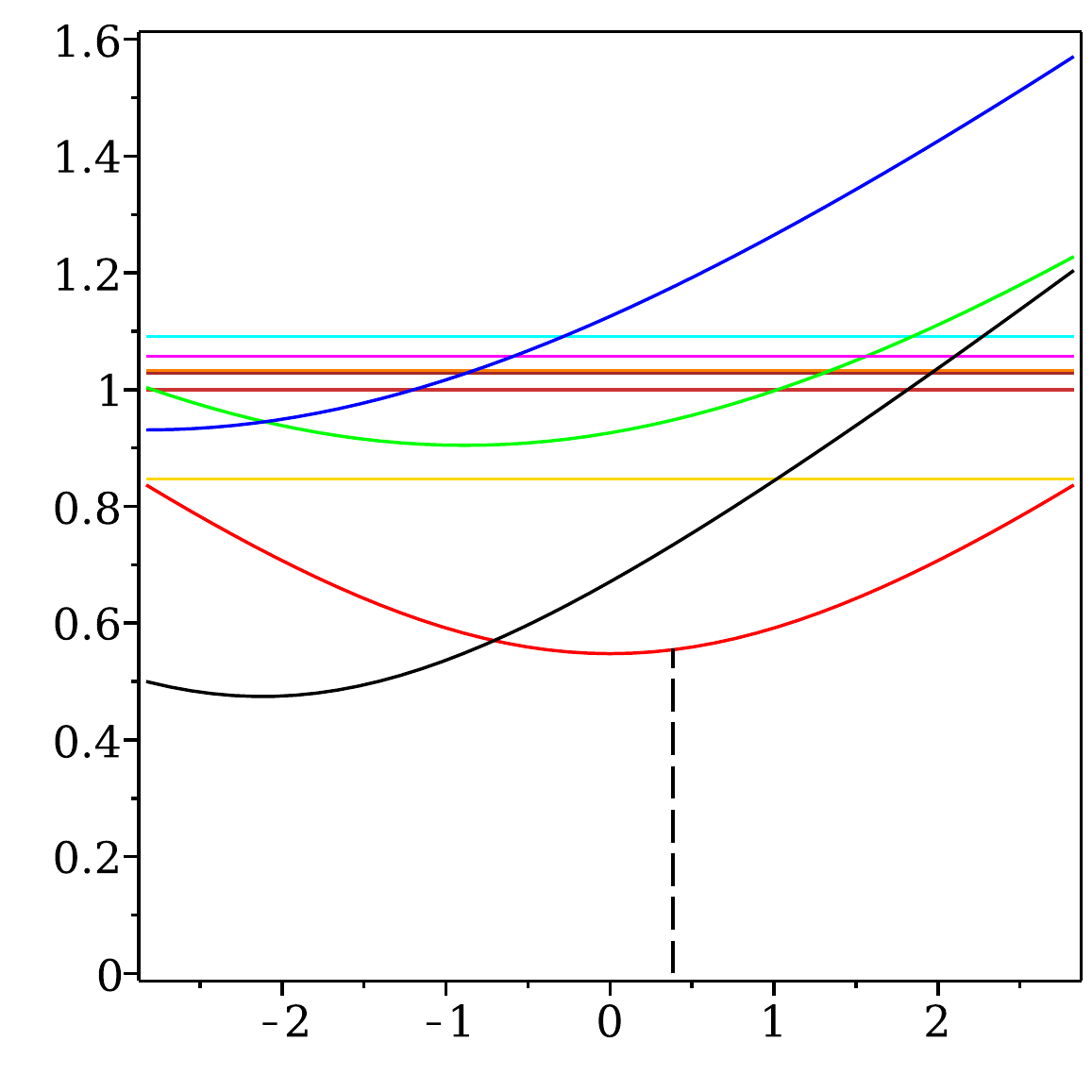}
      \put(45,-6){(i) Case 6}
      \put(0,60){\makebox(0,0){\rotatebox{90}{Distance}}}
        \put(92,3){$t$}
  \end{overpic}
  
    \begin{overpic}
    [width=\linewidth]{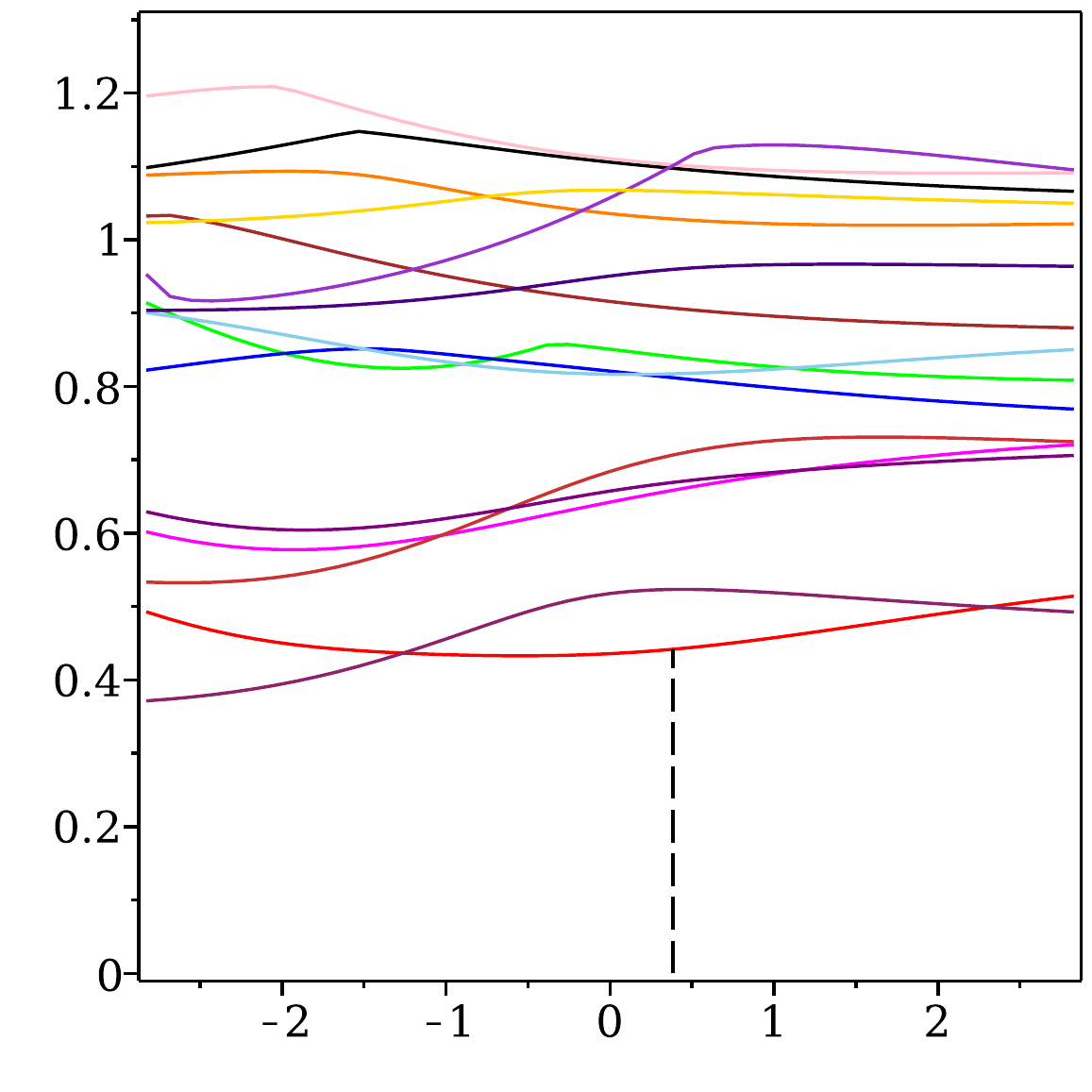}
    \put(45,-6){(j) Case 7}
     \put(0,60){\makebox(0,0){\rotatebox{90}{Distance}}}
        \put(92,3){$t$}
    \end{overpic}
    
    \begin{overpic}
    [width=\linewidth]{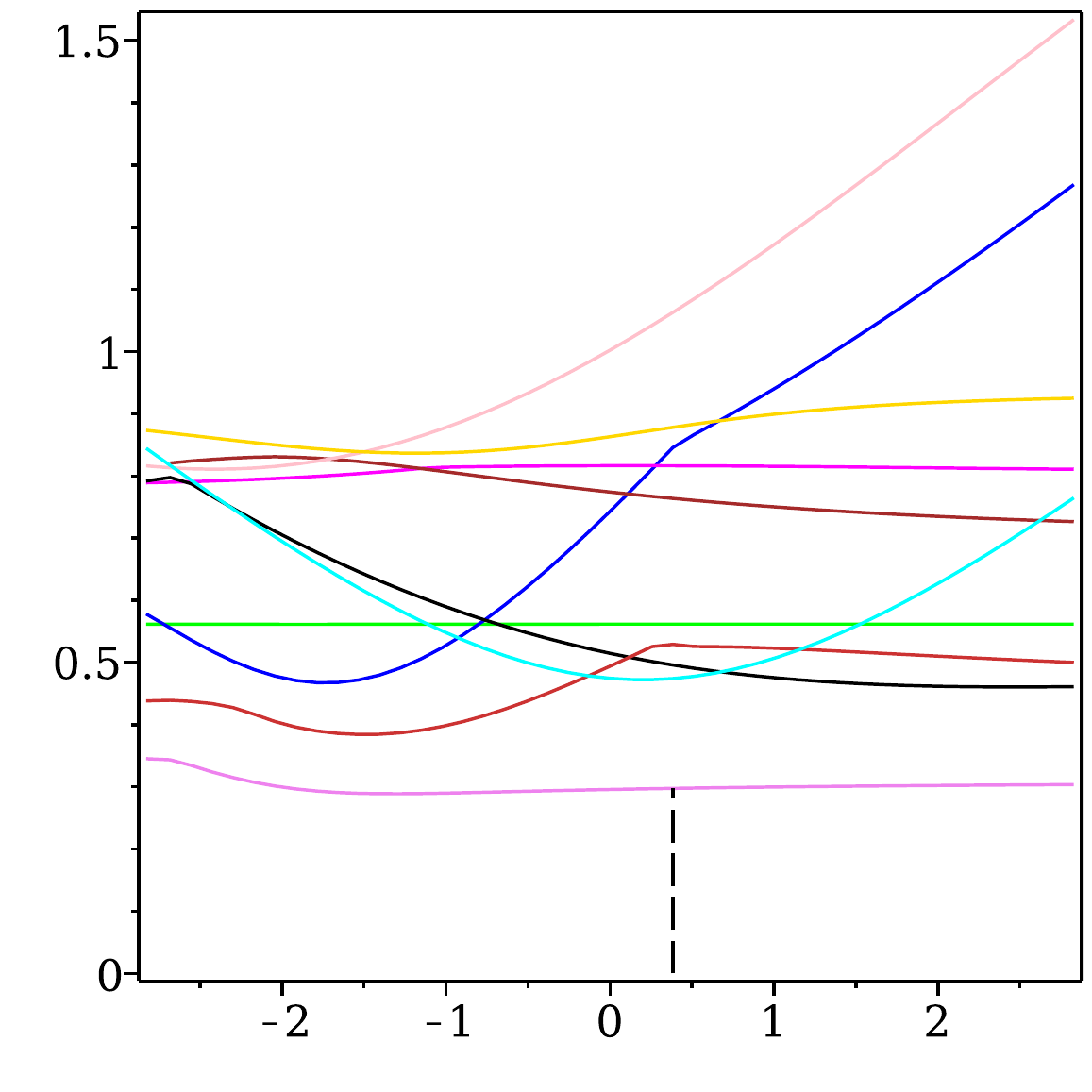}
   \put(45,-6){(k) Case 8}
   \put(0,60){\makebox(0,0){\rotatebox{90}{Distance}}}
        \put(92,3){$t$}
   \begin{small}
   
   \end{small}
  \end{overpic}\par
  
\begin{overpic}
    [width=\linewidth]{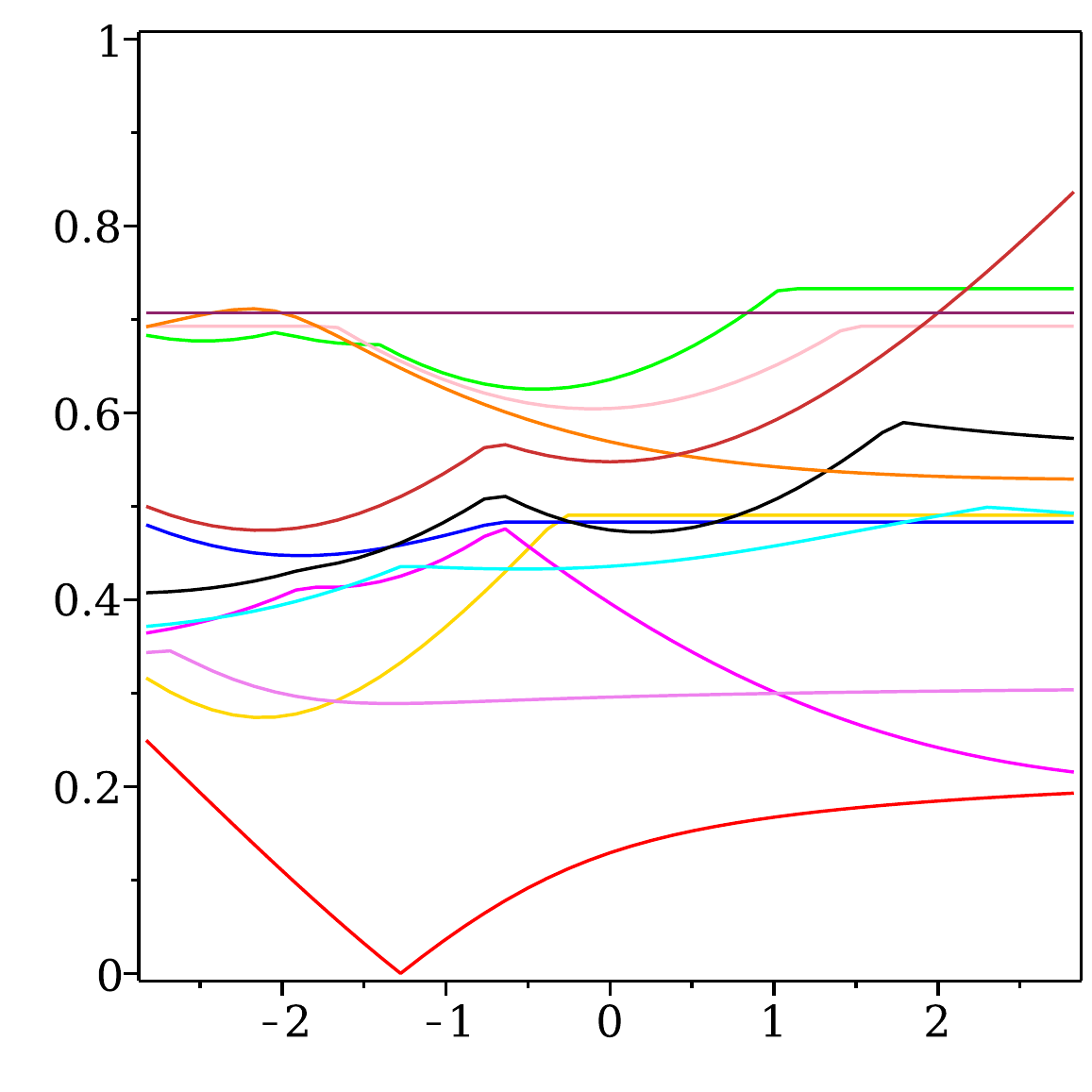}
    \put(45,-6){(l) }
   \put(0,60){\makebox(0,0){\rotatebox{90}{Distance}}}
        \put(92,3){$t$}
 \end{overpic}
 \end{multicols}

\caption{{(a-k) Architectural singularity distance for all the possible design combinations with respect to their architecture singular distance metrics for each case. (l)  The graphs of the minimum overall combinations for each case. 
}} 
\label{graphs}
\end{figure*}
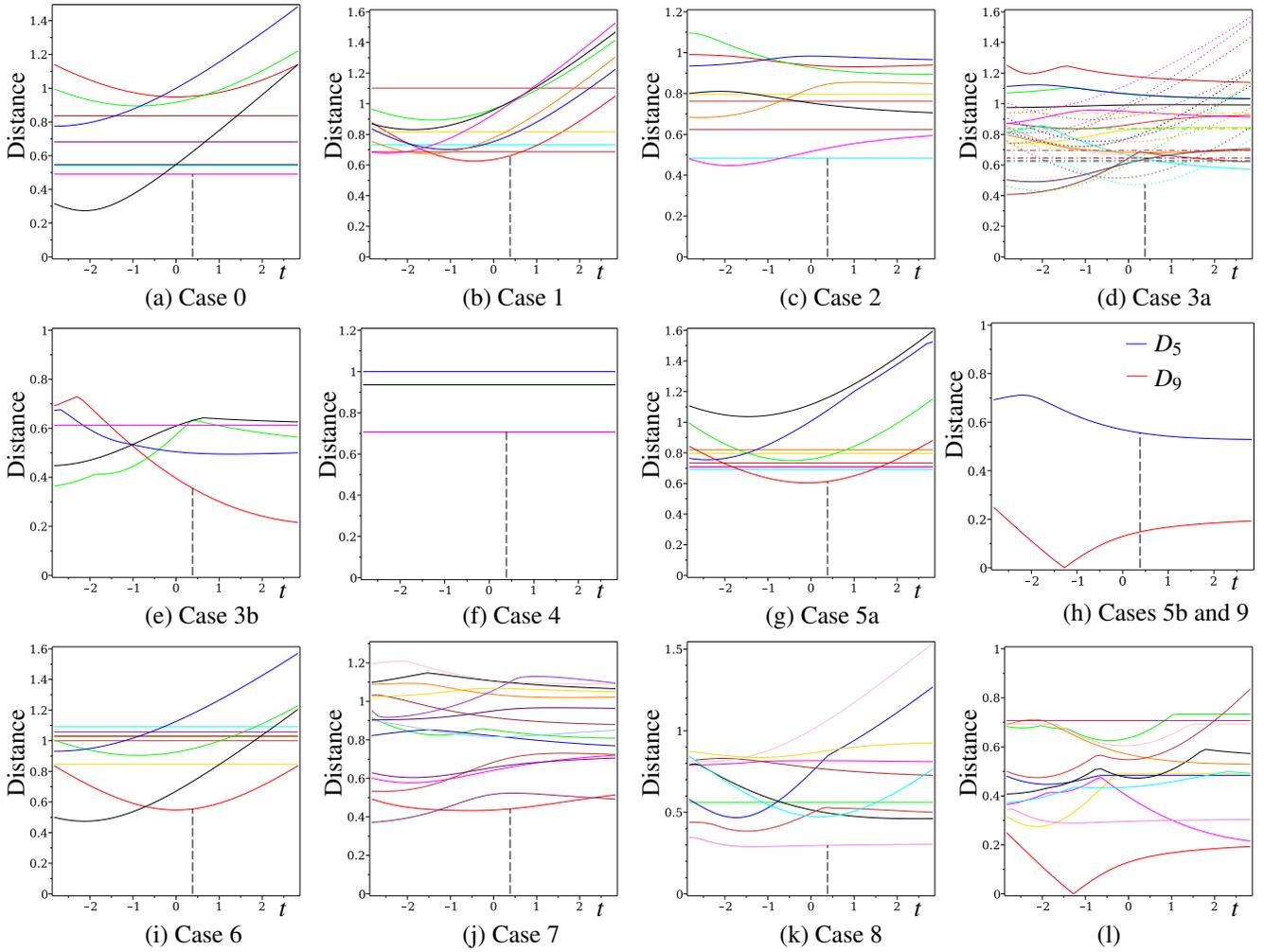

It can be observed in Fig.\ \ref{graphs} that whenever $\mathbf{M}_{5}$ is involved in the design combinations, there is a change in the course of the architecture singularity distance. For the remaining design combinations, this distance remains constant. In Fig.\ \ref{graphs}(a-k), we present for each case the graphs of the minimum distance for each possible combination listed in  Table~\ref{data}. The graphs of the minimum over all combinations for each case are displayed in Fig.\ \ref{graphs}(l). The corresponding architecturally singular designs of these graphs for $t = 0.383206$ are illustrated in 
Fig.\ \ref{geometricdesigns}.
 The coordinates of the base/platform anchor points of these designs are given in Tables \ref{basecoordinates} and \ref{platformcoordinates}. Now, it can be seen from Fig.\  \ref{graphs}(l) that case 9 (red graph) results for all $t$ values in the minimum architecture singularity distance.

\paragraph{Comparison with the already existing method:} To demonstrate the effectiveness of our approach, we compare the obtained architecture singularity distance with the index discussed in Section \ref{sec:review}.

\begin{figure*}[]
\begin{center}
\begin{multicols}{3}
    \begin{overpic}
    [width=50mm]{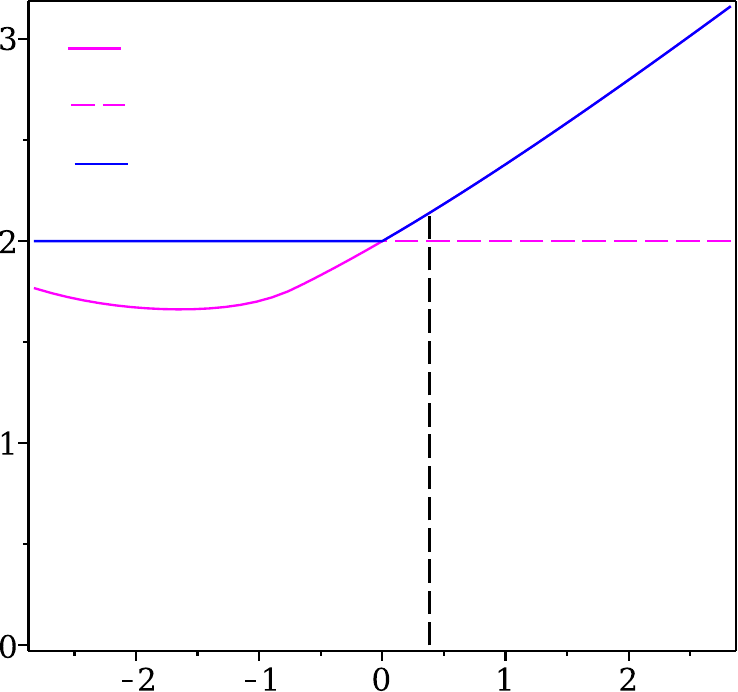}
    \put(45,-6){(a) }
    \put(92,0){$t$}
     \put(-5,60){\makebox(0,0){\rotatebox{90}{radius}}}
     \put(22,86){$\rho_1$}
     \put(22,78){$\rho_2$}
     \put(22,70){max ($\rho_1$, $\rho_2$)}
  \end{overpic}
\begin{overpic}
     [width=50mm]{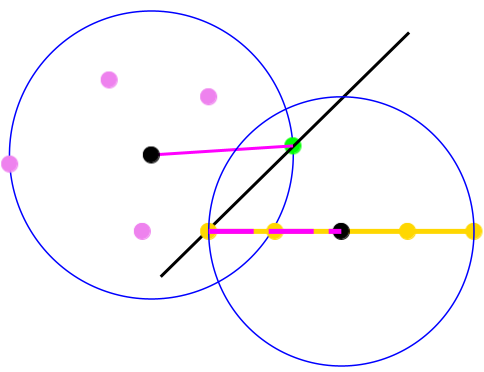}
      \put(45,-25){(b)}
      \put(63,44){$\mathbf{M}_{5}$}
      \put(42,60){$\mathbf{M}_{1}$}
       \put(25,60){$\mathbf{M}_{2}$}
        \put(5,40){$\mathbf{M}_{3}$}
        \put(25,32){$\mathbf{M}_{4}$}
          \put(40,22){$\mathbf{m}_{1}$}
           \put(55,22){$\mathbf{m}_{2}$}
           \put(67,22){$\mathbf{m}_{3}$}
           \put(80,22){$\mathbf{m}_{4}$}
            \put(94,22){$\mathbf{m}_{5}$}
\end{overpic}
 \begin{overpic}
    [width=50mm]{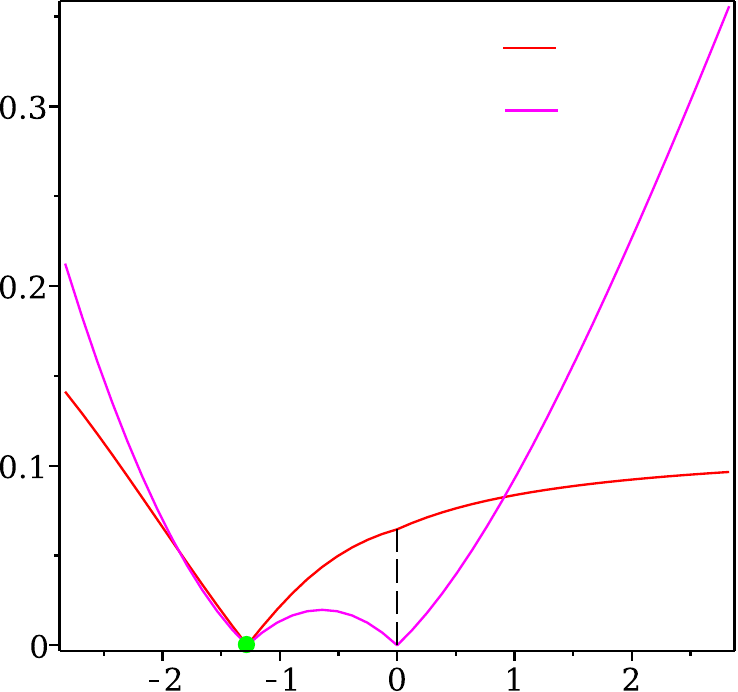}
    \put(45,-6){(c) }
    \put(78,76){$C$}
    \put(78,85) {$D_{9}$}
    \put(92,0){$t$}
     \end{overpic}
\end{multicols}
\end{center}
\caption{{(a) Radius of smallest circle enclosing the base and platform points, respectively. For $t = 0.383206$ these circles are illustrated in (b). (c) Comparison of rescaled architectural singular distance with the index given by Eq.\ (\ref{borrasindex}).}}
\label{comparison}
\end{figure*}
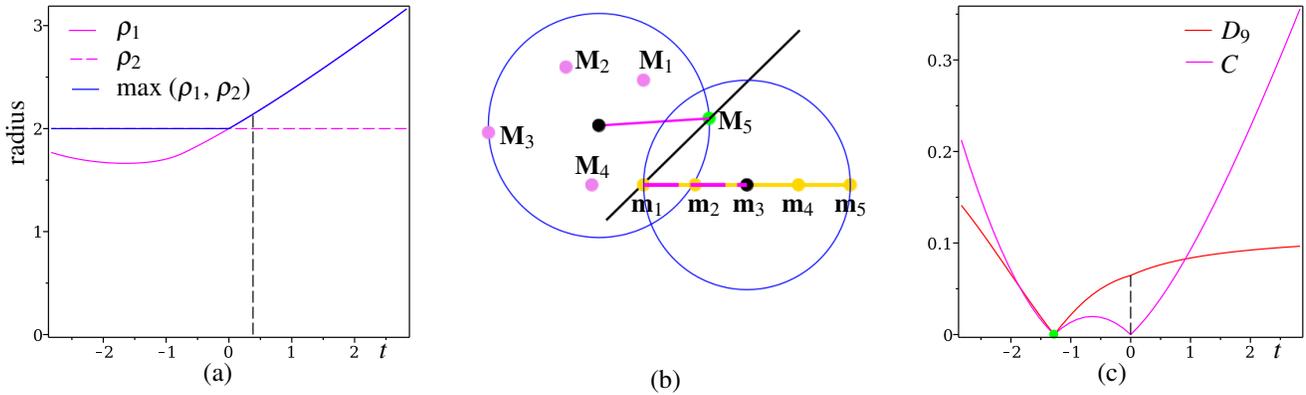

Before doing so, we have to rescale the manipulator in a way that the condition $\max(\rho_{1},\rho_{2})=1$ (cf.\ Section \ref{sec:dist}) holds true. 
 In Fig.\ \ref{comparison}(a), we show the maximum radius of the two circle disks which enclose $\mathbf{M}_{1}, \ldots ,\mathbf{M}_{5}$ and $\mathbf{m}_{1}, \ldots ,\mathbf{m}_{5}$, respectively. 
 For the value $t=0.383206$ these two circles are also illustrated in 
 Fig.\ \ref{comparison}(b).

Finally, in Fig.\ \ref{comparison}(c), we present the rescaled architecture singularity distance, which is compared with the index given by Eq.\ (\ref{borrasindex}) scaled by a factor of $400$ times the original distance.  For $t=-\frac{26302}{20591}$  ($\Longleftrightarrow$  $\mathbf{M}_{5}=(\frac{3}{31},\frac{3}{31})$) the distance is zero, indicating that the design is architecturally singular (indicated with a green dot).
For $t=0$, the leg attachment is coincident with $\mathbf{B}$, which causes the index given by Eq.\ (\ref{borrasindex}) to go to zero,  although the design is not architecturally singular. In contrast, our well-defined architectural singularity distance behaves as expected.

\begin{rmk}
Animations of each case for one of the design combinations showing the closest architectural designs in dependence of $t$ are provided as supplementary materials \cite{b18} for all combinations of a case, which involve $\mathbf{M}_{5}$. In addition, the minimum enclosing circle of the five base anchor points in dependence of $t$ is animated. \hfill $\diamond$
\end{rmk}

\subsection{Example with non-planar base}\label{ex:nonplanar}

For the non-planar base design, we are using the following numerical data as input:
$\mathbf{M}_{1}=(0,0,0)_{\frak{F}_{0}}$, $\mathbf{M}_{2}=(\frac{14}{33},0,0)_{\frak{F}_{0}}$, 
$\mathbf{M}_{3}=(\frac{8}{33}, \frac{4}{33}, 0)_{\frak{F}_{0}}$, $\mathbf{M}_{4}=(\frac{7}{33}, \frac{29}{33}, \frac{32}{33})_{\frak{F}_{0}}$, 
$\mathbf{M}_{5}=(\frac{1}{2}, \frac{-1}{4}, \frac{2}{3})_{\frak{F}_{0}}$, 
$\mathbf{m}_{1}=(0,0,0)_{\frak{F}}$,  $\mathbf{m}_{2}=(\frac{2}{5},0,0)_{\frak{F}}$, $\mathbf{m}_{3}=(1, 0, 0)_{\frak{F}}$, $\mathbf{m}_{4}=(\frac{13}{10},0,0)_{\frak{F}}$, 
$\mathbf{m}_{5}=(\frac{18}{10},0,0)_{\frak{F}}$. Note that this input already fulfills {$max(\rho_{1},\rho_{2})=1$} as  $\rho_2=\tfrac{9}{10}$ holds and the points 
$\mathbf{M}_{1}, \ldots ,\mathbf{M}_{4}$ are located on a unit sphere, which encloses $\mathbf{M}_{5}$. 

The minimum architecture distance for each case $k$ as well as the information of the corresponding combinatorial case encoded in the values for $\ell_{1},\ldots ,\ell_{5}$
are listed in Table \ref{test2}. In this example case 9 yields the global minimizer of the optimization problem,  thus the architecture singular distance equals approximately  
0.0975876652. The 
corresponding architecture singular design is illustrated in Fig.\ \ref{fig:two_bases} and the coordinates of its anchor points are given in 
Appendix~\ref{datacoordinates}.

\begin{figure}[h!]
    \centering
    \begin{subfigure}[b]{0.48\textwidth}
        \begin{overpic}[width=\linewidth]{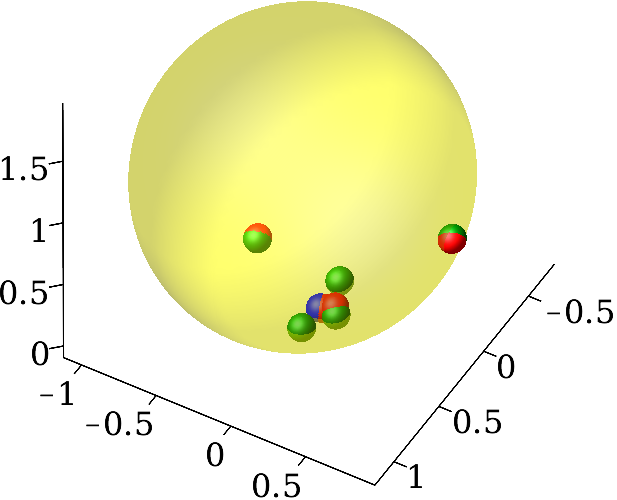}
            \begin{small}
                  \put(52,38.5){$\mathbf{M}_{1}$}
        \put(46,35){\color{blue}$\mathbf{M}^{'}_{1}$}
                \put(41,26){$\mathbf{M}^{'}_{2}$}
                \put(42,31.5){$\mathbf{M}_{2}$}
                \put(57,28){$\mathbf{M}_{3}$}
                \put(57.5,32){$\mathbf{M}^{'}_{3}$}
                \put(65.3,38){$\mathbf{M}^{'}_{4}$}
                \put(70,46.5){$\mathbf{M}_{4}$}
                \put(38,48){$\mathbf{M}^{'}_{5}$}
                \put(34,39){$\mathbf{M}_{5}$}
            \end{small}
        \end{overpic}
        \caption{Base}
    \end{subfigure}
    
    \vspace{1em} 
    \centering
    \begin{subfigure}[b]{0.45\textwidth}
        \begin{overpic}[width=0.9\linewidth]{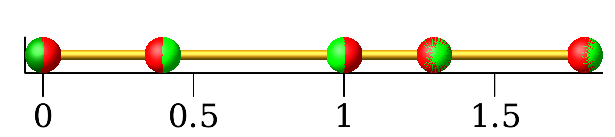}
            \begin{small}
                 \put(2,18){$\mathbf{m}_{1}$}
                 \put(9,16){$\mathbf{m}'_{1}$}
                  \put(20,17){$\mathbf{m}'_{2}$}
                  \put(28,16){$\mathbf{m}_{2}$}
            \put(50,17){$\mathbf{m}_{3}$}
             \put(57,16){$\mathbf{m}'_{3}$}
             \put(64,17){$\mathbf{m}'_{4}$}
             \put(72,16){$\mathbf{m}_{4}$}
              \put(90,18){$\mathbf{m}'_{5}$}
              \put(97,16){$\mathbf{m}_{5}$}
            \end{small}
        \end{overpic}
        \caption{Platform}
    \end{subfigure}
    \caption{The given base (a) and platform (b) anchor points are indicated in green, and the corresponding points of the closest architecture singular design are printed in red. Note that only $\mathbf{M}^{'}_{1}$ is displayed in blue as it is very close to $\mathbf{M}^{'}_{2}$. The corresponding coordinates are given in Appendix~\ref{datacoordinates}.}
    \label{fig:two_bases} 
\end{figure}

\section{Conclusion and future work}\label{future}

We presented an algorithm for computing the closest architecture singular design to a given linear pentapod with respect to an extrinsic metric given in Eq.\ (\ref{eq:distance0}). Another possibility would be to base the computations on an intrinsic metric (e.g.\ the total elastic strain energy density of the platform and base using the physical concept of Green-Lagrange strain \cite{b12}), but this would increase the computational complexity. Even for the 
extrinsic metric under consideration, we are already faced with computational issues as we cannot guarantee the completeness of the solution set for cases 8 and 9 of the minimization problems, which serve as input data for the homotopy continuation approach.

\begin{table}[]
\small\sf\centering
\caption{Computational results obtained for Example \ref{ex:nonplanar}.}
\begin{tabular}{|l||l|l|l|l|l|l|}
\hline
case $k$  & \begin{tabular}[c]{@{}l@{}}minimum $D_k$ \end{tabular} &  $\ell_{1}$  & $\ell_{2}$ &  $\ell_{3}$ &  $\ell_{4}$ &  $\ell_{5}$\\
\hline\hline
0 & 0.1303805266 & 1& 2 &  -- & --  & -- \\ \hline
1 & 0.1001987618 & 1 & 2 & 3 &-- &--\\ \hline
2 & 0.2095878942 & 2 &  3 & 4 & -- &--\\ \hline
3a & 0.1188328328  & 3 & 4&1 &2 &--  \\ \hline
3 & 0.0981384678 & 1 & 2 & 3 & 4 & --  \\ \hline
4 &  0.3205464085  & 2 &3 & 4 &5 &-- \\ \hline
5a &  0.2905620789 & 1 & 5 & 3 & 4 & 2  \\ \hline
5b & 0.3434315159 & 1 & 2 &3 &4 &5\\ \hline
 6 & 0.2183802351 & 1 & 2 & 3 & 4  & 5\\ \hline
 7  & 0.2259492492  &  1 & 2 &  5 & 3 & 4 \\ \hline
 8  & 0.1181118781 & 1   &  2 &  5 & 3 &  4  \\ \hline
 9  & 0.09758766523 &  1 & 2  & 3  & 4 & 5 \\
\hline
\end{tabular}\\[10pt]
\label{test2}
\end{table}

Theoretically, the presented concept can be extended to hexapods (also known as Stewart-Gough platforms) taking into account the classification of their architecturally singular designs given in \cite{b14,b15,b16,b17}, but the arising computational burden prevented us from doing so (even the doubly planar case is out of reach at the moment). On the other end of the scale is its planar analogue; namely the 3-RPR manipulator. 
For this parallel robot the determination of the closest architecture singularity is trivial; besides the possibility that two legs coincide (case 0) there is only one further case that either all three base or platform anchor points collapse into a point, which is just the centroid of the base and platform, respectively.

To pave the way for the presented approach for the applicability to hexapods,  future work should focus on the circumvention of the mentioned computational limitations. 
Moreover, the development of an algorithm for design optimization using the information of the closest architecture singularity is devoted to future research (cf.\ end of Section \ref{sec:dist}). 
We are also interested in finding answers to  the question posted in Remark \ref{rmk:d_D} as well as the missing geometric characterizations of the minimizers of cases 3a, 3b, 5b, 7, 8, and 9, respectively.

\section*{Acknowledgments}
This research is supported by Grant No.\ P 30855-N32 of the Austrian Science Fund FWF. The first author wants to dedicate this article to his friend Jevgeni Igansov on the occasion of his $29^{th}$ birthday.


\appendix      
\section{Appendix} 
\label{datacoordinates} 

The coordinates of the architecture singular design illustrated in Fig.\ \ref{fig:two_bases} are as follows: 
\begin{align*}
\mathbf{M}^{'}_{1}&\approx (0.2298889247, 0.0071714756,-0.0109452093)_{\frak{F}_{0}}, \\ 
\mathbf{M}^{'}_{2}&\approx(0.2257849117, 0.0201911288,-0.0108456958)_{\frak{F}_{0}}, \\
\mathbf{M}^{'}_{3}&\approx(0.2114583372, 0.0933776496, 0.0221428966)_{\frak{F}_{0}}, \\
\mathbf{M}^{'}_{4}&\approx(0.2166543290, 0.8807826507, 0.9680554467)_{\frak{F}_{0}}, \\
\mathbf{M}^{'}_{5}&\approx(0.4950013702,-0.2515229039, 0.6679561979)_{\frak{F}_{0}}, \\
\mathbf{m}^{'}_{1}&\approx(0.0016583117,0,0)_{\frak{F}}, \,\, \mathbf{m}^{'}_{2}\approx(0.3964556801,0,0)_{\frak{F}}, \\ 
\mathbf{m}^{'}_{3}&\approx(1.0026262356,0,0)_{\frak{F}}, \,\,
\mathbf{m}^{'}_{4}\approx(1.2997972749,0,0)_{\frak{F}},  \\
\mathbf{m}^{'}_{5}&\approx(1.7994624976,0,0)_{\frak{F}}.
\end{align*} 

\vspace{-5mm}

\begin{table}[ht!]
\centering
\caption{{Endgame}  settings in \texttt{HC.jl}for cases 8 and 9.}
\begin{tabular}{|l||l|}
\hline
\begin{tabular}[c]{@{}l@{}}\texttt{Endgame}  options \end{tabular} & values \\ \hline \hline
endgame\_start                                                &     0.005    \\ \hline
at\_infinity\_check                                           &      true    \\ \hline
min\_cond                                                    &    1e5     \\ \hline
max\_endgame\_steps                                          &     50\,000    \\ \hline
\end{tabular}
\label{settings1}
\end{table}

\newpage

\begin{table}[ht]
\centering
\caption{Tacker settings in \texttt{HC.jl} for cases 8 and 9.}
\begin{tabular}{|l||l|}
\hline
\begin{tabular}[c]{@{}l@{}}\texttt{Tracker} options \end{tabular} & values \\ \hline \hline
automatic\_differentiation                                               &   3    \\ \hline
extended\_precision                       &      true    \\ \hline
max\_steps                                                    &    60\,000    \\ \hline
min\_setp\_size                                          &     1e-30    \\ \hline
\end{tabular}
\label{settings2}
\end{table}

\vspace{-3mm}

\begin{table}[ht]
\centering
\caption{Comparison of path failures for ab-initio tracking for the planar base design.}
\begin{tabular}{|l||l|l||l|l|}
\hline
 &   \multicolumn{2}{c||} {case 3b (\texttt{HC.jl})} &   \multicolumn{2}{c|} {case 3b (\texttt{Bertini})} \\  
\hline
Run & \multicolumn{1}{l|}{\begin{tabular}[c]{@{}l@{}} $\#$ finite sol-\\utions over $\mathbb{C}$\end{tabular}} & \multicolumn{1}{l||}{\begin{tabular}[c]{@{}l@{}}$\#$ path\\ failures\end{tabular}} & \multicolumn{1}{l|}{\begin{tabular}[c]{@{}l@{}} $\#$ finite sol-\\utions over $\mathbb{C}$\end{tabular}} & \multicolumn{1}{l|}{\begin{tabular}[c]{@{}l@{}}$\#$ path\\ failures\end{tabular}} \\ \hline  \hline
1    & 98             &       42          & 98               & 0             \\ \hline
2    & 98               &    38           & 98               & 0             \\ \hline
3    & 98              &   47             & 98               & 1            \\ \hline
4    & 98               &   55            & 98               & 0             \\ \hline
5    & 98               &     57          & 98               & 0             \\ \hline
\end{tabular}
\label{pathfailures}
\end{table}

\begin{table*}[ht!]
\centering

\caption{Base coordinates for the architectural singular design illustrated in Fig.\ \ref{geometricdesigns}(x) with respective to $\frak{F}_{0}$. Note that a dash indicates $\mathbf{M}'_{i}=\mathbf{M}_{i}$.}
\begin{tabular}{|l||l|l|l|l|l|}
\hline x &
$\mathbf{M}'_{1}$ & $\mathbf{M}'_{2}$  & $\mathbf{M}'_{3}$  & $\mathbf{M}'_{4}$  & $\mathbf{M}'_{5}$  \\ \hline \hline  a &
--  & (-1.25, 1.125) & -- &  (-1.25, 1.125) & -- \\ \hline  b &
(-0.07634, 1.84032) & (-0.07634, 1.84032) & -- & --  & (-0.07634, 1.84032)  \\ \hline  c &
(-0.09250,2.25687) & (-1.34060,1.80740) & (-3.06688, 1.18572) & -- & -- \\ \hline  d &
(0.61872, 1.78748) & (-1.42415,1.56221) & (-3.04231, 1.38378) & -- & (0.61872, 1.78748) \\ \hline  e &
(-0.12856,1.66996) & (-1.32651,1.59984) & (-3.06240,1.49824)  & -- & (1.28844,1.75290) \\ \hline  f &
-- & --  & --  & -- & -- \\ \hline  g &
(0.64685,  1.57211)    &  (-2.98565,0.92010)    & (-1.22197, 1.23667)  & (-1.31511,1.21995) &  (0.64685,  1.57211)   \\ \hline  h &
(0.06306,1.39967)& (-1.39458,1.24654) & (-3.00809, 1.07703)& (-1.13383,1.27393) & (1.24440,1.52378) \\ \hline  i &
(0.63548,1.63548) & -- & -- & -- & (0.63548,1.63548)  \\ \hline  j &
(-0.11639,1.75408) & (-1.43851, 2.37990)& (1.36399, 1.05335) & (-2.94978,0.56046)  & (-1.08833, 0.77315) \\ \hline  k &
(-0.17243, 2.28338) & (-1.25274,1.85109) & (-3.07761,1.12806) & (-1.01627, 0.03120) & (1.29004,1.23440) \\ \hline  l &
(-0.12050, 2.24552) & (-1.33689, 2.00936) & (-3.06684, 1.03951) & (-0.98564, 0.05031)  & (1.28086, 1.17624)  \\ \hline
\end{tabular}
\label{basecoordinates}
\end{table*}

\begin{table*}[ht]
\centering
\caption{Platform coordinates for the architectural singular design illustrated in Fig.\ \ref{geometricdesigns}(x) with respective to $\frak{F}$. Note that a dash indicates $\mathbf{m}'_{i}=\mathbf{m}_{i}$.}
\begin{tabular}{|l||l|l|l|l|l|}
\hline x &
$\mathbf{m}'_{1}$ & $\mathbf{m}'_{2}$  & $\mathbf{m}'_{3}$  & $\mathbf{m}'_{4}$  & $\mathbf{m}'_{5}$   \\ \hline \hline a &
-- & (2,0) & -- & -- & (2,0) \\ \hline b &
(0,0) & (1,0) & (2,0) & (3,0) & (4,0) \\ \hline c &
(1,0) & (1,0) & (1,0) & -- & -- \\ \hline d &
-- & (1.5,0) & (1.5,0) & -- & --  \\ \hline e &
(-0.04618,0) & (1.32906,0) & (1.69933,0) & -- & (4.01778,0) \\ \hline f &
(1.5,0) resp.\ -- & (1.5,0) resp.\ (2.5,0) & (1.5,0) resp.\ (2.5,0) & (1.5,0) resp.\ (2.5,0) & -- resp.\ (2.5,0) \\ \hline g &
(0,0) & (1,0) & (2,0) & (3,0) & (4,0)  \\ \hline h &
(0,0) & (1,0) & (2,0) & (3,0) & (4,0) \\ \hline i &
(2,0) & (2,0) & (2,0) & -- & -- \\ \hline j &
(0.5,0) & (0.5,0) & (2.5,0) & (2.5,0) & --  \\ \hline k &
(-0.03430, 0) & (1.12987, 0) & (1.88461, 0) & (3.50990,0) & (3.50990,0) \\ \hline l &
(-0.05688,0) & (1.15110, 0) & (1.96204, 0) & (2.89201, 0) & (4.05172, 0) \\ \hline
\end{tabular}
\label{platformcoordinates}
\end{table*}

\end{document}